\documentclass[letterpaper]{article} 
\usepackage{aaai23}  
\usepackage{times}  
\usepackage{helvet}  
\usepackage{courier}  
\usepackage[hyphens]{url}  
\usepackage{graphicx} 
\usepackage{natbib}  
\usepackage{caption} 
\usepackage{algorithm}
\usepackage{algorithmic}

\usepackage{newfloat}
\usepackage{listings}
\DeclareCaptionStyle{ruled}{labelfont=normalfont,labelsep=colon,strut=off} 
\floatstyle{ruled}
\newfloat{listing}{tb}{lst}{}
\floatname{listing}{Listing}
\title{Towards Optimal Randomized Strategies in Adversarial Example Game}

\author{Jiahao Xie\textsuperscript{\rm 1},
        Chao Zhang\thanks{Corresponding author.}\textsuperscript{\rm 2},
        Weijie Liu\textsuperscript{\rm 3,1},
        Wensong Bai\textsuperscript{\rm 1,2},
        Hui Qian\textsuperscript{\rm 1,4}\\
}
\affiliations{
\textsuperscript{\rm 1}College of Computer Science and Technology, Zhejiang University\\
\textsuperscript{\rm 2}Advanced Technology Institute, Zhejiang University \\
\textsuperscript{\rm 3}Qiushi Academy for Advanced Studies, Zhejiang University \\
\textsuperscript{\rm 4}State Key Lab of CAD\&CG, Zhejiang University\\
xiejh@zju.edu.cn, zczju@zju.edu.cn, westonhunter@zju.edu.cn, wensongb@zju.edu.cn, qianhui@zju.edu.cn
}

\usepackage{mathtools}
\usepackage{lingmacros}
\usepackage{tree-dvips}
\usepackage{amsmath}
\usepackage{amssymb}
\usepackage{amsthm}
\usepackage{xcolor}
\usepackage{etoolbox}
\usepackage{booktabs}
\usepackage{subfiles}
\usepackage{threeparttable}
\usepackage{multirow}
\usepackage{nicefrac}
\usepackage{enumerate}
\usepackage[shortlabels]{enumitem}
\newcommand{\SUB}[1]{\ENSURE \textbf{#1}}

\newcommand{\algorithmicdoinparallel}{\textbf{do in parallel}}
\makeatletter
\AtBeginEnvironment{algorithmic}{%
  \newcommand{\FORALLP}[2][default]{\ALC@it\algorithmicforall\ #2\ %
    \algorithmicdoinparallel\ALC@com{#1}\begin{ALC@for}}%
}
\makeatother
\usepackage{accents}

\usepackage[labelformat=simple]{subcaption}
\newcommand{\bftab}{\fontseries{b}\selectfont}
\input{required.tex}
\newcommand{\AlgName}{Fully Randomized Adversarial Training}
\newcommand{\AlgAbbr}{FRAT}

\allowdisplaybreaks

\usepackage{eqparbox}

\makeatletter
\patchcmd{\algorithmic}{\addtolength{\ALC@tlm}{\leftmargin} }{\addtolength{\ALC@tlm}{\leftmargin}}{}{}
\makeatother

\begin{document}
\maketitle

\begin{abstract}
The vulnerability of deep neural network models to adversarial example attacks is a practical challenge in many artificial intelligence applications. 
A recent line of work shows that the use of randomization in adversarial training is the key to find optimal strategies against adversarial example attacks. 
However,  in a fully randomized setting where both the defender and the attacker can use randomized strategies, there are no efficient algorithm for finding such an optimal strategy.
To fill the gap, we propose the first algorithm of its kind,  called FRAT,  which models the problem with a new infinite-dimensional continuous-time flow on probability distribution spaces. 
FRAT maintains a lightweight mixture of models for the defender,  with flexibility to efficiently update mixing weights and model parameters at each iteration. 
Furthermore, FRAT utilizes lightweight sampling subroutines to construct a random strategy for the attacker.
We prove that the continuous-time limit of FRAT converges to a mixed Nash equilibria in a zero-sum game formed by a defender and an attacker. 
Experimental results also demonstrate the efficiency of FRAT on CIFAR-10 and CIFAR-100 datasets.
\end{abstract}


\section{Introduction}

Deep Neural Network (DNN) models have been shown to be highly vulnerable to adversarial example attacks, which are tiny and imperceptible perturbations of the input designed to fool the model~\cite{biggio2013evasion, szegedy2014intriguing, goodfellow2015explaining}. 
The vulnerability severely hindered the use of DNNs in safety-critical applications and became one of the main concerns of the artificial intelligence community~\cite{goodfellow2015explaining}. 
To improve the robustness of the model to adversarial examples, various defense strategies have been proposed in the past few years~\cite{goodfellow2015explaining, papernot2016distillation, samangouei2018defense, madry2018towards, cohen2019certified, moosavi2019robustness, zhang2019theoretically, pinot2020randomization, meunier2021mixed, gowal2021improving}. 
Among existing strategies, the adversarial training (AT) approach~\cite{goodfellow2015explaining, madry2018towards} is widely recognized as the most successful one~\cite{schott2018towards, pang2020bag, maini2020adversarial, bai2021recent}, which usually constructs a virtual attacker that generates the worst adversarial examples maximizing the loss in the neighborhood of clean examples and seeks a robust classification model (classifier) by minimizing the loss on the generated examples.

Recent studies in the AT literature begin to explore randomized strategies for classifiers that probabilistically mix multiple classification models and show that stochastic classifiers are more robust to adversarial examples than a single deterministic classifier~\cite{xie2018mitigating, wang2019resnets, pinot2019theoretical, pinot2020randomization, meunier2021mixed}. 
\citet{pinot2020randomization} demonstrate from a game-theoretic perspective that randomized classifiers provide better worst-case theoretical guarantees than deterministic ones when attackers use deterministic strategies.
They empirically show that the mixture of two classifiers obtained by their proposed Boosted Adversarial Training (BAT) algorithm achieves significant improvement over the state-of-the-art deterministic classifier produced by the Standard Adversarial Training (SAT) algorithm~\cite{madry2018towards}. 
Later, \citet{meunier2021mixed} show that randomized classifiers also outperform deterministic ones when attackers are allowed to use sophisticated randomized attack strategies.

In particular, \citet{meunier2021mixed} established the existence of an Mixed Nash Equilibrium (MNE) in randomized adversarial training games, i.e., there is an optimal randomized strategy pair of classifier and attacker such that neither of them can benefit from unilaterally changing its strategy, whereas when the player uses a deterministic strategy, a Nash equilibrium may not exist.
For problems with discrete classifier parameter spaces, \citet{meunier2021mixed} propose two theoretically-guaranteed algorithms and then heuristically extend them to problems with continuous parameter. However, their heuristics are not guaranteed to find an MNE, and the efficiency may decrease as the number of mixture components in the randomized classifier increases (see the results in our experiments).

In this paper, we propose an efficient algorithm named Fully Randomized Adversarial Training (FRAT) for finding MNE in the randomized AT game with continuous classifier parameter spaces. 
In particular, FRAT maintains a weighted mixture of classification models for the classifier, where both model parameters and mixture weights are updated in each round with relatively low computational cost.
Furthermore, adversarial examples for the randomized attacker are generated by a restricted sampling subroutine called Projected Langevin algorithm (PLA), which has similar computations to Projected Gradient Descent (PGD) used in existing AT algorithms.
Note that the actions of the classifier and the attacker in each round are actually obtained through a continuous-time stream of discrete game objective functions that converge to the MNE of a randomized AT game. Our main contributions are summarized as follows.
\begin{enumerate}[1.]
\item 
We propose a new hybrid continuous-time flow that appropriately exploits the bilinear problem structure of the randomized AT game. For the classifier, we adopt the Wasserstein-Fisher-Rao (WFR) flow of the objective because it leads to fast convergence of the objective in the probability space, and efficient update rules for model parameters and mixture weights can be achieved by discretizing this flow with the first order Euler scheme; For the attacker, we first construct a surrogate function by adding a regularizer to the bilinear game, since it is often difficult to construct a proper flow for the inner constrained maximization problem in the original game. Using this surrogate, a convergent Logit Best Response (LBR) flow can be derived, whose first order Euler discretization can be efficiently computed by PLA.

\item
We develop analyses for the proposed hybrid continuous-time flow. With a proper regularization parameter, this flow is proved to converge at an $\OM(1/T)$ rate to an MNE of the original unregularized AT game under mild assumptions, where $T$ denotes the time.
\end{enumerate}

We conduct numerical experiments on synthetic and real datasets to compare the performance of the proposed algorithm and existing ones. Experimental results demonstrate the efficiency of the proposed algorithm.


\section{Related Works}

\paragraph{Randomized strategies in adversarial training.}
Several works have investigated randomized strategies in adversarial training~\cite{bulo2016randomized, perdomo2019robust, bose2020adversarial, pinot2020randomization, meunier2021mixed}.
Notably, \citet{pinot2020randomization} prove, from a game theoretical point of view, that randomized classifiers offer better worst-case theoretical guarantees than deterministic ones when the attacker uses deterministic strategies.
\citet{meunier2021mixed} further show the existence of MNE in the adversarial example game when both the classifier and the attacker use randomized strategies.
Existing methods for finding randomized strategies can be divided into two classes: (i) noise injection methods~\cite{xie2018mitigating, dhillon2018stochastic, wang2019resnets} and (ii) mixed strategy methods~\cite{bulo2016randomized, perdomo2019robust, bose2020adversarial, pinot2020randomization, meunier2021mixed}, both of which suffer from critical limitations.
The first class of methods inject random noise into the input data or certain layers of the classification model, which is shown to be effective in practice but generally lacks theoretical guarantees.
The second class of methods construct randomized strategies in probability spaces using game theory, and are usually theoretically-guaranteed.
However, existing methods of this class apply to restricted settings where the randomized strategies are restricted in certain families of distributions~\cite{bulo2016randomized, bose2020adversarial, pinot2020randomization}, or the strategy spaces are discrete and finite~\cite{perdomo2019robust, meunier2021mixed}.

\paragraph{Algorithms for finding MNE in zero-sum games.}
In recent years, there has been an increasing interest in finding mixed Nash equilibria in two-player zero-sum continuous games~\cite{perkins2014stochastic, hsieh2019finding, suggala2020follow, domingo2020mean, liu2021infinite, ma2021provably}.
However, these algorithms are impractical or infeasible for the adversarial example game.
Specifically, the algorithms in~\cite{hsieh2019finding, domingo2020mean, liu2021infinite, ma2021provably} only apply to games on unconstrained spaces or manifolds, and do not apply to the adversarial example game in which the strategy space of the attacker is a compact convex constraint set.
Although \citet{perkins2014stochastic} and~\citet{suggala2020follow} develop algorithms that apply to zero-sum games with compact convex strategy spaces, their algorithms need store all historical strategies of both the players during the optimization process, which is prohibitive for the adversarial example game with medium to large-scale datasets.


\section{Problem Setting and Algorithm}  \label{section_algorithm}

\subsection{Problem Setting}
The literature of adversarial training usually formulate the problem of adversarial example defense/attack as an Adversarial Training (AT) game, i.e., a zero-sum game between the classifier and the attacker~\cite{shaham2018understanding, madry2018towards}.
Suppose that we are given a dataset $\DM = \{(\xB_i, y_i)\}_{i=1}^N$, where $(\xB_i, y_i) \in \XM \times \YM$ denotes the feature-label pair of the $i$-th data sample, a classification model parameterized by $\theta \in \Theta$, and a loss function $\ell: \Theta \times (\XM \times \YM) \rightarrow \RBB$.
The attacker seeks strong adversarial examples by perturbing sample within a given distance $\varepsilon$ to maximize the loss function $\ell$, while the classifier aims to learn a model that minimizes the loss function defined on the generated adversarial data samples.
Specifically, the deterministic AT game is given by
\begin{equation}  \label{eq_finite_problem}
    \inf_{\theta \in \Theta} {\textstyle \frac{1}{N} \! \sum_{i=1}^N} \Big( \sup_{\hat{\xB}_i \in \BBB_{\varepsilon}(\xB_i)} \! \ell(\theta, (\hat{\xB}_i, y_i)) \Big),
\end{equation}
where $\BBB_{\varepsilon}(\xB_i) := \{\xB \in \XM : d(\xB, \xB_i) \le \varepsilon\}$ and $d(\cdot, \cdot)$ denotes the distance function on $\XM$.
Instead of searching deterministic strategies as in~\eqref{eq_finite_problem}, we consider a randomized setting of adversarial training, where the classifier (resp., the attacker) searches randomized strategies in the space $\MM_1^+(\Theta)$ (resp., $\MM_1^+(\BBB_{\varepsilon}(\xB_i)), i \in \{1, \ldots, N\}$).
Here, $\MM_1^+(\Theta)$ (resp., $\MM_1^+(\BBB_{\varepsilon}(\xB_i))$) denotes the Polish space of Borel probability measures on $\Theta$ (resp., $\BBB_{\varepsilon}(\xB_i)$).
Note that a randomized strategy of the attacker can be written as $\nu := (\nu_1, \ldots, \nu_N) \in \Sigma$, where $\Sigma$ stands for the product space $\MM_1^+(\BBB_{\varepsilon}(\xB_1)) \times \ldots \times \MM_1^+(\BBB_{\varepsilon}(\xB_N))$.
Then, the randomized AT game can be formulated as the following infinite-dimensional minimax problem on the probability space
\begin{equation}  \label{eq_infinite_problem}
    \inf_{\mu \in \MM_1^+(\Theta)} \sup_{\nu \in \Sigma} \Big\{ \! \LM(\mu, \nu) \!:=\! {\textstyle \frac{1}{N} \! \sum_{i=1}^N} \EBB_{\theta \sim \mu, \xB \sim \nu_i} [\ell(\theta, (\xB, y_i))] \Big\}
\end{equation}
\citet{meunier2021mixed} prove that under mild assumptions, there exists a mixed Nash equilibrium in~\eqref{eq_infinite_problem}, that is, there is a pair of strategy $(\mu^*, \nu^*)$ such that, for any $\mu \in \MM_1^+(\Theta)$ and $\nu \in \Sigma$,
$
    \LM(\mu^*, \nu) \le \LM(\mu^*, \nu^*) \le \LM(\mu, \nu^*)
$.
In contrast, the deterministic formulation~\eqref{eq_finite_problem} does not always have a pure Nash equilibrium~\cite{pinot2020randomization}.
This necessitates the use of randomized strategies for finding Nash equilibria (see~\cite{meunier2021mixed} for more discussions).

\paragraph{Entropy regularization.} Instead of directly solving~\eqref{eq_infinite_problem}, we add an entropy regularization term to make the objective function strongly concave in $\nu$.
Note that this is a common technique in the infinite-dimensional optimization literature~\cite{perkins2014stochastic, domingo2020mean, ma2021provably, meunier2021mixed}.
Specifically, we define the regularization function as $\HM(\nu) := \frac{1}{N} \sum_{i=1}^N \textrm{KL}(\nu_i \| u_i)$, where $u_i$ is the uniform distribution over $\BBB_{\varepsilon}(\xB_i)$ and $\textrm{KL}(\cdot \| \cdot)$ denotes the Kullback-Leibler (KL) divergence, i.e., $\textrm{KL}(\nu_i \| u_i) = \int \log (\frac{d \nu_i}{d u_i}) d \nu_i, \ $ if $\nu_i$ is absolutely continuous w.r.t.\ $u_i$, otherwise $\textrm{KL}(\nu_i \| u_i) = +\infty$.
With this regularization function, we define the regularized adversarial example game as
\begin{equation}  \label{eq_regularized_problem}
    {\textstyle \inf_{\mu \in \MM_1^+(\Theta)} \sup_{\nu \in \Sigma}} \left\{ \LM(\mu, \nu) - \beta \HM(\nu) \right\},
\end{equation}
where $\beta > 0$ is the regularization parameter.

\subsection{The Proposed Algorithm}
To solve~\eqref{eq_regularized_problem}, we propose an algorithm named {\AlgName} ({\AlgAbbr}).
{\AlgAbbr} is derived by discretizing a continuous-time flow on the probability spaces $\MM_1^+(\Theta)$ and $\Sigma$.
Constructing a continuous-time flow (defined by an Ordinary/Partial Differential Equation (ODE/PDE)) and then discretizing it to obtain an algorithm is a common routine for optimization on the probability space~\cite{welling2011bayesian, liu2017stein, liutkus2019sliced, domingo2020mean, ma2021provably}.
Note that even for this complicated infinite-dimensional space, it is feasible to analyze the convergence of a continuous-time flow using various ODE/PDE analysis tools.
With a well-behaved flow, a practical and efficient discrete-time algorithm can be naturally derived using standard discretization techniques.
This routine has also been widely used to obtain algorithms with good convergence properties for optimization on $\RBB^d$ (see~\cite{su2014differential} and references therein).
In what follows, we first propose a hybrid continuous-time flow and then derive the {\AlgAbbr} algorithm.

\subsubsection{The Continuous-time Flow.}
Here, we construct a hybrid continuous-time flow of $(\mu(t), \bar{\nu}(t)) \in \MM_1^+(\Theta) \times \Sigma$ that guarantees descent on the space $\MM_1^+(\Theta)$ and ascent on $\Sigma$, where $\mu(t)$ and $\bar{\nu}(t)$ denote the strategies of the classifier and the attacker, respectively.

We let the strategy $\mu(t)$ of the classifier follow the Wasserstein-Fisher-Rao (WFR) flow
\begin{align}  \label{eq_flow_classifier}
    \dot{\mu}(t)
    ={}& \gamma \nabla \cdot \left( {\textstyle \frac{\mu(t)}{N} \sum_{i=1}^N} \EBB_{\xB \sim \nu_i(t)}[\nabla_{\theta} \ell(\theta, (\xB, y_i))]\right)  \nonumber \\
        &+ \alpha \mu(t) \left( \LM(\mu(t), \nu(t)) - \LM(\delta_{\theta}, \nu(t)) \right),
\end{align}
with the initial condition $\mu(0) = \mu^0$ for some $\mu^0 \in \MM_1^+(\Theta)$, where $\gamma$ and $\alpha$ are non-negative constants and $\nu(t) = (\nu_1(t), \ldots, \nu_N(t))$ will be defined later.
Actually, the WFR flow~\eqref{eq_flow_classifier} is the gradient flow of the objective function $\LM$ on the Wasserstein-Fisher-Rao space, and $\LM$ descends following this flow when the strategy $\nu(t)$ is kept fixed.
Note that the WFR flow has been widely used in optimization problems on the probability space, such as over-parameterized network training~\cite{liero2018optimal, rotskoff2019global, chizat2022sparse} and unconstrained randomized zero-sum games~\cite{domingo2020mean}.

The attacker uses the Logit Best Response (LBR) flow
\begin{align}  \label{eq_flow_attacker}
    \dot{\bar{\nu}}(t) = {\textstyle \frac{1}{t}} \left( \nu(t) - \bar{\nu}(t) \right)
\end{align}
with the initial condition $\bar{\nu}(t) = \nu^0$ for $t \in [0, 1]$.
Here, $\nu(t)$ is the best response to the time-average strategy $\bar{\mu}(t) := \frac{1}{t} \int_{0}^{t} \mu(s) d s$ of the classifier for $t \ge 1$, i.e.,
\begin{align}  \label{eq_nu_t}
    \nu(t) := 
    \begin{cases}
        \nu^{0}, &\textrm{if} \ t \in [0, 1) \\
        \underset{\nu \in \Sigma}{\textrm{argmax}} \ \LM ( \bar{\mu}(t), \nu ) \!-\! \beta \HM(\nu), &\textrm{otherwise}.
    \end{cases}
\end{align}
The LBR flow has been widely used in constrained games, where the minimization/maximization sub-problem is defined on constrained sets~\cite{hofbauer2002global, perkins2014stochastic, lahkar2015logit}.
As we shall see in the next section, \eqref{eq_nu_t} results in increasing of  certain potential function and induces convergence to MNE when combined with~\eqref{eq_flow_classifier}.
We call the hybrid flow of $(\mu(t), \bar{\nu}(t))$ following~\eqref{eq_flow_classifier}-\eqref{eq_nu_t} as the WFR-LBR flow.

\subsubsection{The Discrete-time Algorithm.}
To obtain a practical algorithm, we discretize the WFR-LBR flow in both space and time.
The discretization steps are detailed as follows.
\begin{enumerate}[1.]
\item Discretization of the WFR flow~\eqref{eq_flow_classifier}.
First, we use a weighted mixture $\hat{\mu}(t) := \sum_{j=1}^M w_j(t) \delta_{\theta_{j}(t)} \in \MM_1^+(\Theta)$ to approximate $\mu(t)$ defined on the whole space, where $M$ is a fixed integer, $\delta_{\theta_j(t)}$ is the Dirac measure of mass $1$ at the particle $\theta_j(t)$, $w_j(t) \ge 0$ is the mixing weight such that $\sum_{j=1}^M w_j(t) = 1$.
Then, we construct the following continuous-time flow for $\theta_j(t)$ and $w_j(t)$
\begin{align}  \label{eq_dynamics_particles_classifier}
\resizebox{.88\linewidth}{!}{$
\begin{cases}
    \dot{\theta}_j(t) = \!-\! \frac{\gamma}{N} \sum_{i=1}^N \EBB_{\xB \sim \nu_i(t)}[\nabla_{\theta} \ell(\theta_j(t), (\xB, y_i))] \\ 
    \dot{w}_j(t) \!=\! \alpha \! \left(\LM(\mu(t), \nu(t)) \!-\! \LM(\delta_{\theta_j(t)}, \nu(t)) \right) w_j(t).
\end{cases}
$}
\end{align}
Note that~\eqref{eq_flow_classifier} is actually derived from~\eqref{eq_dynamics_particles_classifier} and the mean field limit of~\eqref{eq_dynamics_particles_classifier} converges to~\eqref{eq_flow_classifier}~\cite{domingo2020mean}.
By applying the first order Euler discretization to~\eqref{eq_dynamics_particles_classifier}, we obtain the following update rule 
\begin{align}  \label{eq_update_rule_particles_classifier}
\resizebox{.88\linewidth}{!}{$
\begin{cases}
    \theta_j^{(t+1)} \!=\! \theta_j^{(t)} \!-\! \frac{\eta}{N}\sum_{i=1}^{N} \nabla_{\theta} \ell(\theta_j^{(t)}, (\hat{\xB}_i^{(t)}, y_i)) \\
    w_j^{(t+1)} \!=\! \frac{w_j^{(t)} \exp(- \frac{\eta'}{N}\sum_{i=1}^{N} \ell(\theta_j^{(t)}, (\hat{\xB}_i^{(t)}, y_i)))}{\sum_{j=1}^M w_j^{(t)} \exp(- \frac{\eta'}{N}\sum_{i=1}^{N} \ell(\theta_j^{(t)}, (\hat{\xB}_i^{(t)}, y_i)))},
\end{cases}
$}
\end{align}
where $\eta$ and $\eta'$ are positive step sizes and the superscript $(t)$ denotes the discrete time step.
This update rule moves each particle $\theta_j(t)$ along the negative gradient direction to decrease the loss value and adjusts the weights so that particles with lower loss values have larger weights.

\item Discretization of the LBR flow of~\eqref{eq_flow_attacker}.
The first order Euler discretization of~\eqref{eq_flow_attacker} leads to 
\begin{align}  
    \bar{\nu}_i^{(t+1)} \leftarrow {\textstyle \frac{t+1}{t+2}} \bar{\nu}_i^{(t)} + {\textstyle \frac{1}{t+2}} \nu_i^{(t+1)},
\end{align}
As both $\bar{\nu}$ and $\nu_i$ are infinite-dimensional variables, it is generally hard to compute the above Euler discretization.
It can be verified that the maximization problem in~\eqref{eq_nu_t} has a unique solution $\nu(t)$, and each $\nu_i(t)$ has the density
\begin{equation}  \label{eq_density_logit_best_response}
\resizebox{.88\linewidth}{!}{$
    p_{\nu_i}(\xB)
    = \frac{\exp(\beta^{-1} \EBB_{\theta \sim \bar{\mu}(t)}[\ell(\theta, (\xB, y_i))])}{\int_{\BBB_{\varepsilon}(\xB_i)} \exp(\beta^{-1} \EBB_{\theta \sim \bar{\mu}(t)}[\ell(\theta, (\xB, y_i))]) d \xB}.
$}
\end{equation}
Thus, we can use the stochastic approximation technique to approximate $\nu_i(t)$ by drawing a sample from $\nu_i^{(t+1)}$~\eqref{eq_density_logit_best_response}, and obtain the following update rule
\begin{align}  \label{eq_update_rule_attacker}
    \bar{\nu}_i^{(t+1)} \leftarrow {\textstyle \frac{t+1}{t+2}} \bar{\nu}_i^{(t)} + {\textstyle \frac{1}{t+2}} \delta_{\hat{\xB}_i^{(t+1)}},
\end{align}
where the initial distribution $\bar{\nu}_i^{(0)} = \delta_{\hat{\xB}_i^{(0)}}$ for some $\hat{\xB}_i^{(0)} \in \BBB_{\varepsilon}(\xB_i)$, and $\hat{\xB}_i^{(t+1)}$ is sampled from~\eqref{eq_density_logit_best_response}.
Note that the objective function in~\eqref{eq_nu_t} is the sum of a linear function and a nonlinear regularization function.
Thereby, the update rule of $\bar{\nu}$ can be viewed as the Generalized Frank-Wolfe (aka Generalized Conditional Gradient) algorithm~\citep{bonesky2007generalized, bredies2009generalized} on the probability measure space.
In addition, the update rule of $\bar{\nu}$ is an extension of the stochastic fictitious play~\citep{hofbauer2002global, perkins2014stochastic} from the one-dimensional space to a high-dimensional space.
\end{enumerate}
By combining the update rules~\eqref{eq_update_rule_particles_classifier} and~\eqref{eq_update_rule_attacker}, we obtain the {\AlgAbbr} algorithm, which is summarized in Algorithm~\ref{algorithm_wfr_gfw}.
On line 1 of Algorithm~\ref{algorithm_wfr_gfw}, the strategies of the classifier and the attacker are initialized.
Lines 3-6 computes the update direction of the classifier's strategy.
Specifically, line 3 (resp., lines 4-5) computes model parameters (resp., weights) of the classifier's update direction, and line 6 combines the weights and models to obtain the update direction.
Line 7 constructs the attacker's update direction by sampling from the distribution in~\eqref{eq_density_logit_best_response}.
Finally, the two players update their strategies on lines 8 and 9, respectively.

To sample from the distribution $\nu_i^{(t)}$ which is supported on the constraint set $\BBB_{\varepsilon}(\xB_i)$, we resort to an efficient constrained sampling method called the Projected Langevin Algorithm (PLA)~\cite{bubeck2018sampling}.
PLA produces an approximate sample by performing the following update step for multiple iterations
\begin{align}  \label{eq_projectd_langevin_update}
\resizebox{.88\linewidth}{!}{$
\hat{\xB}_{i} \!\leftarrow\!
    \Pi_{\BBB_{\varepsilon(\xB_i)}} \! (\hat{\xB}_i \!+\! {\textstyle \frac{\lambda}{2 \beta}} \EBB_{\theta \sim \bar{\mu}^{(t)}} \nabla_{\xB} \ell(\theta, (\hat{\xB}_i, y_i))
    \!+\! \omega \sqrt{\lambda} \xi),
$}
\end{align}
where $\Pi_{\BBB_{\varepsilon(\xB_i)}} (\cdot)$ denotes the projection onto $\BBB_{\varepsilon(\xB_i)}$, $\lambda$ is the step size, $\omega$ is a constant, and $\xi$ is sampled in each iteration from $\NM(\zeroB, \IB)$, i.e., the Gaussian distribution with mean $\zeroB$ and covariance matrix $\IB$.
Actually, the PLA algorithm can be viewed as a variant of Projected Gradient Descent (PGD) with the only addition of Gaussian noise perturbations.
Thus, the computational cost of PLA is similar to PGD which is used in existing AT algorithms.

\begin{algorithm}[t]
\caption{{\AlgName}.}
\label{algorithm_wfr_gfw}
\begin{algorithmic}[1]
\SUB{Input:}
IID samples $\theta_1^{(0)}, \ldots, \theta_M^{(0)}$ from $\mu^{(0)} \in \MM_1^+(\Theta)$,
an IID sample $\hat{\xB}_i^{(0)}$ from $\nu_i^{(0)} \in \MM_1^+(\BBB_{\varepsilon}(\xB_i))$ for each $i \in \{1, \ldots, N\}$,
initial weights $w_j^{(0)} = 1/M$ for $j \in \{1, \ldots, M\}$,
and step sizes $\eta$ and $\eta'$.

\STATE $\bar{\mu}^{(0)} \leftarrow \sum_{j=1}^M w_j^{(0)} \delta_{\theta_j^{(0)}}$, $\bar{\nu}_i^{(0)} \leftarrow \delta_{\hat{\xB}_i^{(0)}}$ for each $i \in \{1, \ldots, N\}$;

\FOR{$t = 0, \ldots, T - 1$}
    \STATE $\theta_j^{(t+1)} \leftarrow \theta_j^{(t)} - \frac{\eta}{N}\sum_{i=1}^{N} \nabla_{\theta} \ell(\theta_j^{(t)}, (\hat{\xB}_i^{(t)}, y_i))$ for each $j \in \{1, \ldots, M\}$;

    \STATE $\hat{w}_j^{(t+1)} \leftarrow w_j^{(t)} \exp(- \frac{\eta'}{N}\sum_{i=1}^{N} \ell(\theta_j^{(t)}, (\hat{\xB}_i^{(t)}, y_i)))$ for each $j \in \{1, \ldots, M\}$;

    \STATE $w_j^{(t+1)} \leftarrow \hat{w}_j^{(t+1)} / \sum_{j=1}^M \hat{w}_j^{(t+1)}$ for each $j \in \{1, \ldots, M\}$;

    \STATE $\mu^{(t+1)} \leftarrow \sum_{j=1}^M w_j^{(t+1)} \delta_{\theta_j^{(t+1)}}$;

    \STATE Sample $\hat{\xB}_i^{(t+1)}$ from $\nu_i^{(t+1)}$ defined in~\eqref{eq_density_logit_best_response} for each $i \in \{1, \ldots, N\}$;   \label{line_sampling_subroutine}

    \STATE $\bar{\mu}^{(t+1)} \leftarrow \frac{t+1}{t+2} \bar{\mu}^{(t)} + \frac{1}{t+2} \mu^{(t+1)}$;

    \STATE $\bar{\nu}_i^{(t+1)} \leftarrow \frac{t+1}{t+2} \bar{\nu}_i^{(t)} + \frac{1}{t+2} \delta_{\hat{\xB}_i^{(t+1)}}$ for each $i \in \{1, \ldots, N\}$;
\ENDFOR

\SUB{Return} $\bar{\mu}^{(T)}$ and $\{\bar{\nu}_i^{(T)}\}_{i=1}^N$.

\end{algorithmic}
\end{algorithm}

In what follows, we discuss some tricks and techniques to speed up the proposed algorithm.
\begin{enumerate}[1.]
    \item
    In practice, it is not necessary to sample $\{\hat{\xB}_1^{(t)}, \ldots, \hat{\xB}_N^{(t)}\}$ in step 7 and store them for calculating $\bar{\nu}^{(t)}$ in step 9 as in a defense task, the defender is only concerned with finding an optimal classifier, and thus the output strategy of the attacker is negligible.
    We only need to sample $\hat{\xB}_i^{(t)}$'s in step 3 and 4 according to ~\eqref{eq_density_logit_best_response} with $\bar{\mu}^{(t)}$.
    This greatly reduces the storage cost.
    \item The second technique is to use stochastic minibatch gradients to approximate the full gradient.
    Specifically, in the update steps of $\theta_j^{(t)}$ and $w_j^{(t)}$, we can sample a minibatch $\BM_{\xB}$ of the perturbed data samples from $\{\hat{\xB}_1^{(t)}, \ldots, \hat{\xB}_N^{(t)}\}$ to estimate the exact loss value $\frac{1}{N}\sum_{i=1}^{N} \ell(\theta_j^{(t)}, (\hat{\xB}_i^{(t)}, y_i))$ and its gradient.
    In this way, we can perform the sampling subroutine on line~\ref{line_sampling_subroutine} of Algorithm~\ref{algorithm_wfr_gfw} for only a minibatch of data points in each iteration because the rest are unused.
    \item In each step~\eqref{eq_projectd_langevin_update} of PLA, we can also randomly select a minibatch $\BM_{\mu}$ from $\{\mu^{(0)}, \ldots, \mu^{(t)}\}$ to approximate the full gradient $\EBB_{\theta \sim \bar{\mu}^{(t)}}[\nabla \ell(\theta, (\xB, y_i))]$.
    Furthermore, to reduce the space complexity, we can maintain a sliding window $\{\mu^{(t - |\BM_{\mu}| + 1)}, \ldots, \mu^{(t)}\}$ and compute $\frac{1}{|\BM_{\mu}|} \sum_{s=t-|\BM_{\mu}|+1}^{t} \EBB_{\theta \sim \mu^{(s)}}[\nabla \ell(\theta, (\xB, y_i))]$ as a surrogate of $\EBB_{\theta \sim \bar{\mu}^{(t)}}[\nabla \ell(\theta, (\xB, y_i))]$.
\end{enumerate}
By using the above techniques and an $S$-step PLA algorithm as the sampling subroutine, {\AlgAbbr} computes $(M |\BM_{\xB}| + M S |\BM_{\mu}| |\BM_{\xB}| )$ gradients in each iteration, where the computation over $\BM_{\mu}$ can be parallelized.
In comparison, the SAT algorithm~\cite{madry2018towards} with $S$ steps of PGD attack in each iteration computes $(S + 1) |\BM_{\xB}|$ gradients.
Thus, when $M$ and $|\BM_{\mu}|$ are small, our algorithm has a similar per-iteration computation time to SAT.


\section{Analysis}
In this section, we show that the continuous-time flow~\eqref{eq_flow_classifier} and~\eqref{eq_flow_attacker} converges to an approximate MNE.
Here, we only present the major results and defer the detailed analyses to Appendix~\ref{proofs}.
Throughout our analysis, the following three assumptions are required.
\begin{assumption}  \label{assumption_loss_function}
    The loss function $\ell: \Theta \times (\XM \times \YM) \rightarrow \RBB$ satisfies the following conditions: 
        (i) the function $\ell$ is non-negative and Borel measurable;
        (ii) $\ell(\theta, (\xB, y))$ is continuous differentiable and $G$-Lipschitz w.r.t.\ $(\theta, \xB)$;
        (iii) $\exists U > 0$, $\forall \theta \in \Theta$ and $(\xB, y) \in \XM \times \YM$, $0 \le \ell(\theta, (\xB, y)) \le U$.
\end{assumption}

\begin{assumption}  \label{assumption_primal_space}
    $\Theta$ is a compact Riemannian manifold without boundary of dimension $d_{\theta}$ embedded in $\RBB^{D_{\theta}}$.
    For all $\theta \in \Theta$, $\textrm{Vol}(\BBB_{\varepsilon}(\theta)) \ge e^{-K} \varepsilon^{d_{\theta}}$, where the volume is defined in terms of the Borel measure of $\Theta$.
\end{assumption}

\begin{assumption}  \label{assumption_initial_value}
    The initial distribution $\mu^{0}$ of the classifier has a Radon-Nikodym derivative $\rho : = \frac{d \mu^{0}}{d \theta}$ with respect to the Borel measure of $\Theta$ and $\rho(\theta) \ge e^{-K'}$ for all $\theta \in \Theta$.
    Similarly, $\nu_i^{0}$ also has a Radon-Nikodym derivative $q_i : = \frac{d \nu_i^{0}}{d \xB}$ with respect to the Lebesgue measure of $\BBB_{\varepsilon}(\xB_i)$ for all $i \in \{1, \ldots, N\}$ and $q_i(\xB) > 0$ for all $\xB \in \BBB_{\varepsilon}(\xB_i)$.
\end{assumption}

Under Assumptions~\ref{assumption_loss_function} and~\ref{assumption_primal_space}, the infimum and supremum in both the problems~\eqref{eq_infinite_problem} and~\eqref{eq_regularized_problem} can be achieved, and there exists MNE in these problems.
We refer the reader to Appendix~\ref{section_existence_of_mne} for details.
A natural metric for measuring the quality of a candidate solution $(\tilde{\mu}, \tilde{\nu})$ of the regularized problem~\eqref{eq_regularized_problem} is the primal-dual gap
\begin{align}  \label{eq_primal_dual_gap}
    \GM_{\beta}(\tilde{\mu}, \tilde{\nu})
    :={}& \sup_{\nu \in \times_{i=1}^N \MM_1^+(\BBB_{\varepsilon}(\xB_i))} \{\LM(\tilde{\mu}, \nu) - \beta \HM(\nu)\}  \nonumber \\
        &- \inf_{\mu \in \MM_1^+(\Theta)} \{\LM(\mu(T), \tilde{\nu}) - \beta \HM(\tilde{\nu})\}.
\end{align}
Note that when $\GM_{\beta}(\tilde{\mu}, \tilde{\nu}) = 0$, $(\tilde{\mu}, \tilde{\nu})$ is an MNE of~\eqref{eq_regularized_problem}.
In the case $\beta = 0$, $\GM_{\beta}$ is the primal-dual gap for the problem~\eqref{eq_infinite_problem}.

\subsection{Regularization Error Analysis}
In the following theorem, we provide an upper bound of the approximation error due to the entropy regularization.
\begin{theorem}  \label{theorem_regularization_error}
    Let $\XM$ be $\RBB^{d_{\xB}}$ equipped with the $\ell_{\infty}$ norm and let $\BBB_{\varepsilon}(\xB_i) = \{\xB \in \XM: \| \xB - \xB_i \|_{\infty}\} \le \varepsilon$.
    Under Assumption~\ref{assumption_loss_function} and the condition $0 < \beta \le \varepsilon / d_{\xB}$, we have
    \begin{align*}
        \GM_0(\tilde{\mu}, \tilde{\nu})
        \le \GM_{\beta}(\tilde{\mu}, \tilde{\nu})
            + \beta d_{\xB} \log {\textstyle \frac{2 \varepsilon G}{\beta d_{\xB}}}
            + \beta d_{\xB}.
    \end{align*}
\end{theorem}
Theorem~\ref{theorem_regularization_error} shows that an (approximate) MNE of~\eqref{eq_regularized_problem} is still an approximate MNE of~\eqref{eq_infinite_problem} when the regularization level $\beta$ is sufficiently small.

\subsection{Convergence Analysis}
In what follows, we analyze the convergence rate of the proposed WFR-LBR dynamics.
By the definition of $\nu(T)$ in~\eqref{eq_nu_t},
when $T \ge 1$, the primal-dual gap of $(\bar{\mu}(T), \bar{\nu}(T))$ produced by the WFR-LBR flow can be written as
\begin{align}  \label{eq_primal_dual_gap_T_ge_1}
    &\GM_{\beta}(\bar{\mu}(T), \bar{\nu}(T))  \nonumber
    ={} \LM(\bar{\mu}(T), \nu(T)) - \beta \HM(\nu(T))  \nonumber \\
        &- \inf_{\mu \in \MM_1^+(\Theta)} \{\LM(\mu(T), \bar{\nu}(T)) - \beta \HM(\bar{\nu}(T))\},
\end{align}
We construct two new potential functions $\RM_{\mu}(T)$ and $\RM_{\nu}(T)$ that allow us to separately analyzing the convergence of the classifier and the attacker
\begin{align}  \label{eq_Lyapunov_functions}
\resizebox{.88\linewidth}{!}{$
\begin{cases}
    \RM_{\mu}(T)
    :={}& \hspace{-1em} \frac{1}{T} \int_0^T \LM(\mu(t), \nu(t)) d t
        - \LM(\mu^*(T), \bar{\nu}(T)) \\
    \RM_{\nu}(T)
    :={}& \hspace{-1em} \LM(\bar{\mu}(T), \nu(T))
        - \beta \HM(\nu(T)) \\
        & \hspace{-1em} - \frac{1}{T} \int_0^T \LM(\mu(t), \nu(t)) d t
        + \beta \HM(\bar{\nu}(T)),
\end{cases}
$}
\end{align}
where $\mu^*(T)$ is defined as
$
    \mu^*(T) \in \textrm{argmin}_{\mu \in \MM_1^+(\Theta)} \LM(\mu, \bar{\nu}(T)).
$
Note that $\GM_{\beta}(\bar{\mu}(T), \bar{\nu}(T)) = \RM_{\mu}(T) + \RM_{\nu}(T)$ when $T \ge 1$.
Therefore, to analyze the convergence of $\GM_{\beta}(\bar{\mu}(T), \bar{\nu}(T))$, it suffices to bound the two potential functions.
The upper bounds of $\RM_{\mu}(T)$ and $\RM_{\nu}(T)$ are provided in Lemmas~\ref{lemma_primal_Lyapunov} and~\ref{lemma_dual_Lyapunov} below, respectively.

\begin{lemma}  \label{lemma_primal_Lyapunov}
    Under Assumptions~\ref{assumption_loss_function}-\ref{assumption_initial_value},
    $
        \RM_{\mu}(T)
        \le \frac{1}{\alpha T} (K + K' + d_{\theta} (1 - \log d_{\theta} + \log(\alpha (U + G) T)))
            + \frac{\gamma}{2} (U + G)^2 T.
    $
\end{lemma}

By Lemma~\ref{lemma_primal_Lyapunov}, with sufficiently small $\gamma$ and large enough $\alpha$ and $T$, we can bound $\RM_{\mu}(T)$ by any desired accuracy.

\begin{lemma}  \label{lemma_dual_Lyapunov}
    Under Assumptions~\ref{assumption_loss_function}-\ref{assumption_initial_value}, $\forall T \ge 1$, we have
    $
        \RM_{\nu}(T) \le {\textstyle \frac{1}{T}} \RM_{\nu}(1),
    $
    where
    $
    \RM_{\nu}(1)
    = \max_{\nu \in \Sigma}
        \{
            \LM(\bar{\mu}(1), \nu) - \beta \HM(\nu)
        \}  \allowbreak
        - (\LM(\bar{\mu}(1), \nu^{0}) - \beta \HM(\nu^{0}))
    \ge 0
    $.
\end{lemma}
Lemma~\ref{lemma_dual_Lyapunov} indicates that the potential function $\RM_{\nu}(T)$ decreases at an $\OM(1/T)$ rate.
By combining Lemmas~\ref{lemma_primal_Lyapunov} and~\ref{lemma_dual_Lyapunov}, we obtain the following convergence result.
\begin{theorem}  \label{theorem_primal_dual_gap}
Under Assumptions~\ref{assumption_loss_function}-\ref{assumption_initial_value}, for any $T \ge 1$, we have
$
    \GM_{\beta}(\bar{\mu}(T), \bar{\nu}(T))
    \le \frac{1}{\alpha T} (K + K' + d_{\theta} (1 - \log d_{\theta} + \log(\alpha (U + G) T)))
        + \frac{\gamma}{2} (U + G)^2 T
        + \frac{1}{T} \RM_{\nu}(1).
$
\end{theorem}
Theorem~\ref{theorem_primal_dual_gap} shows that if we set $\alpha = \OM(1)$ and $\gamma = \OM(1/T^2)$, $(\bar{\mu}(T), \bar{\nu}(T))$ is an $\tilde{\OM}(1/T)$-approximate mixed Nash equilibrium of the regularized game~\eqref{eq_regularized_problem}.
Further, under the condition of Theorem~\ref{theorem_regularization_error} and the additional condition that $\beta \le \OM(1/T)$, $(\bar{\mu}(T), \bar{\nu}(T))$ is also an $\tilde{\OM}(1/T)$-approximate mixed Nash equilibrium of the original game~\eqref{eq_infinite_problem}.


\section{Numerical Experiments}  \label{section_experiments}

\begin{figure}
\begin{subfigure}{.49\linewidth}
  \centering
  \includegraphics[width=\linewidth]{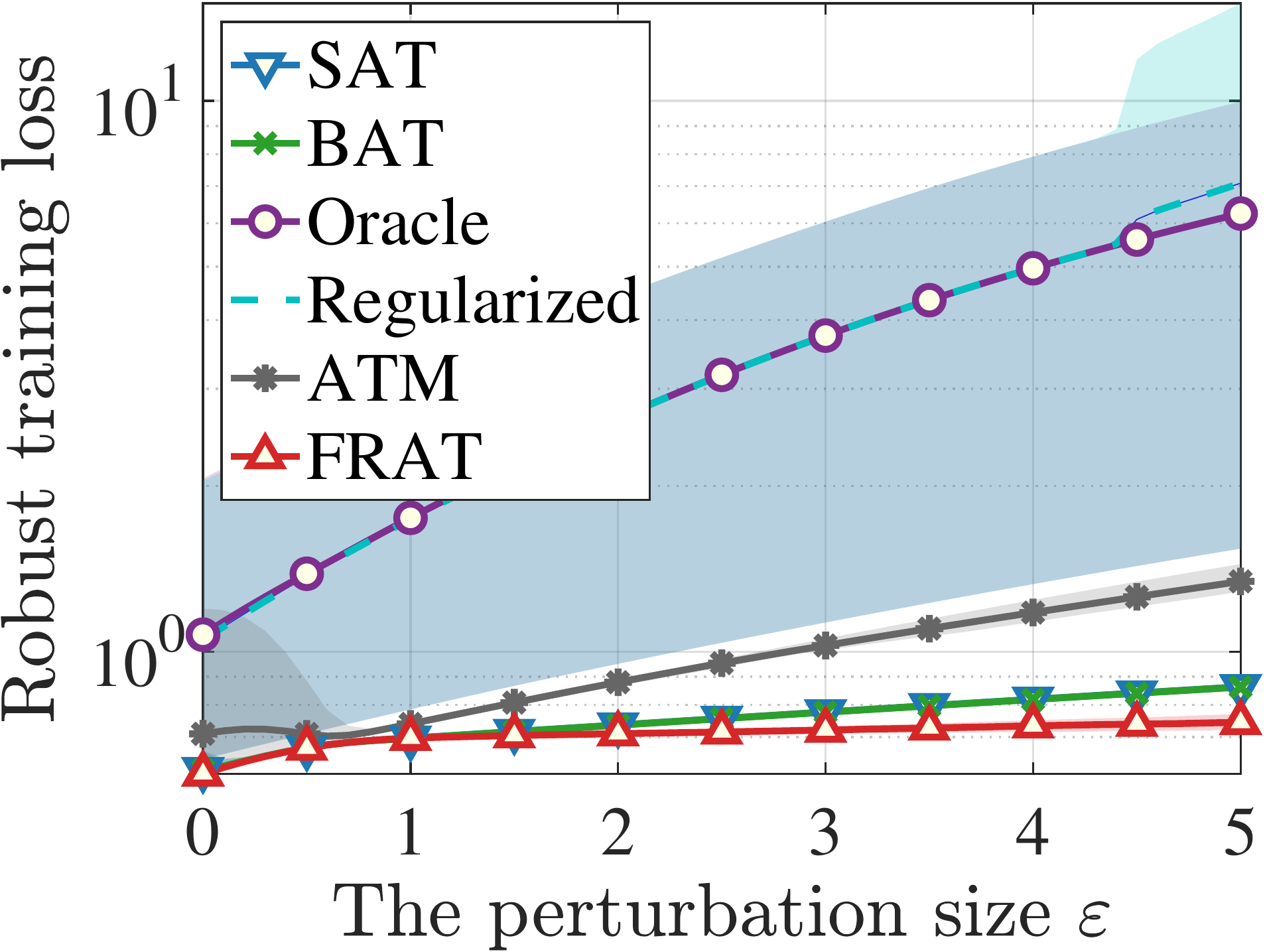}
\end{subfigure}
\hfill
\begin{subfigure}{.49\linewidth}
  \centering
  \includegraphics[width=\linewidth]{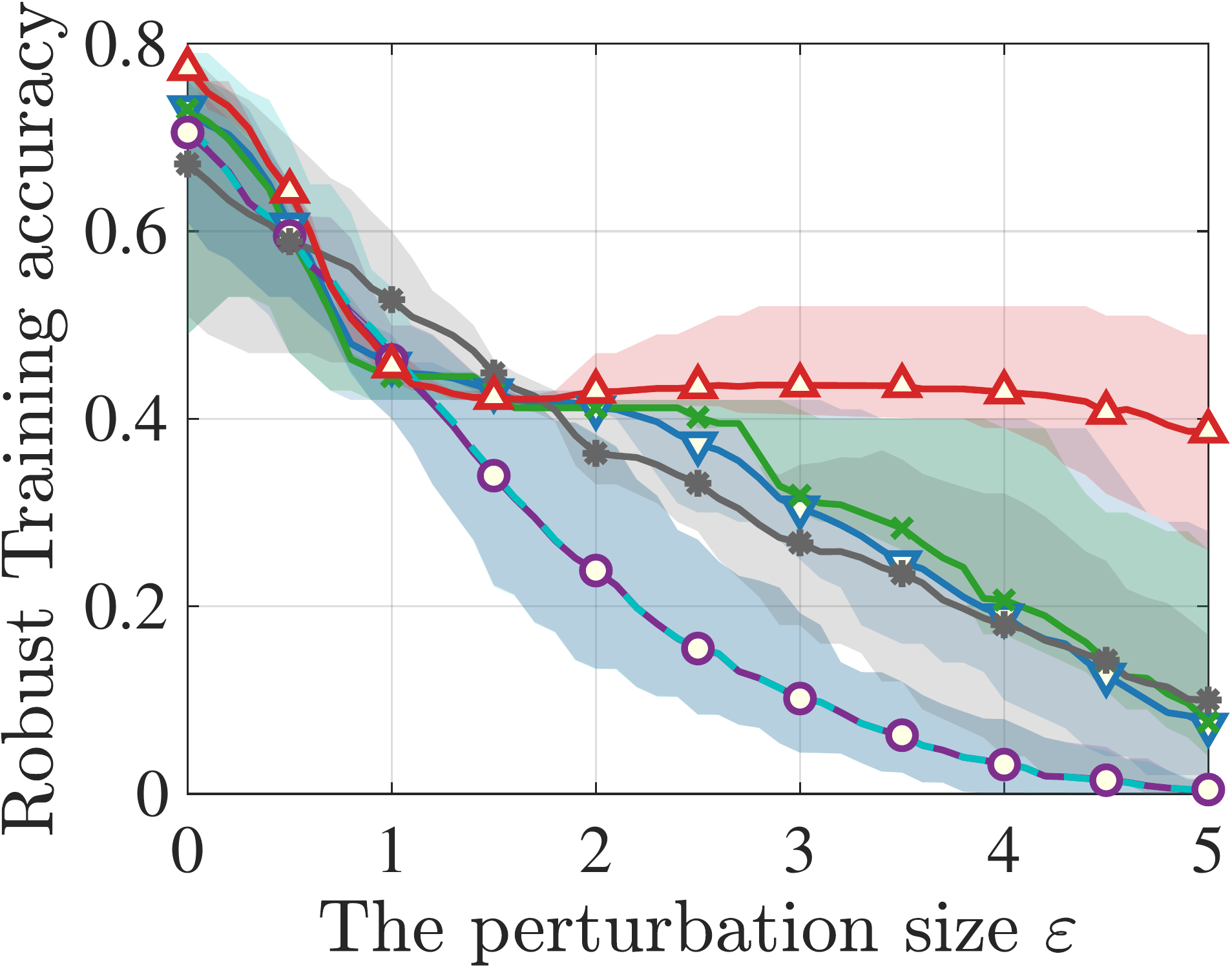}
\end{subfigure}
\vspace*{.5em}
\\
\begin{subfigure}{.49\linewidth}
  \centering
  \includegraphics[width=\linewidth]{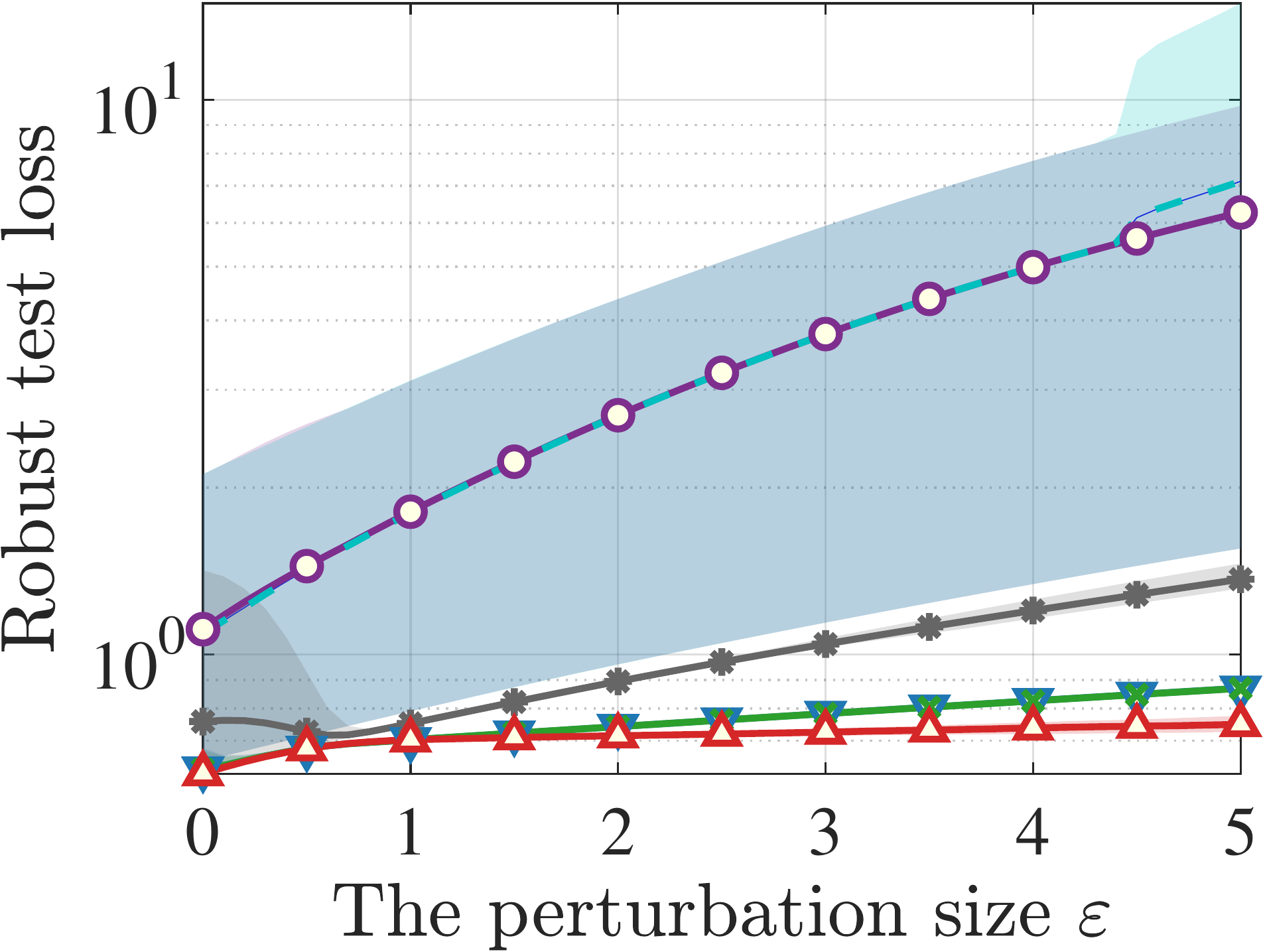}
\end{subfigure}
\hfill
\begin{subfigure}{.49\linewidth}
  \centering
  \includegraphics[width=\linewidth]{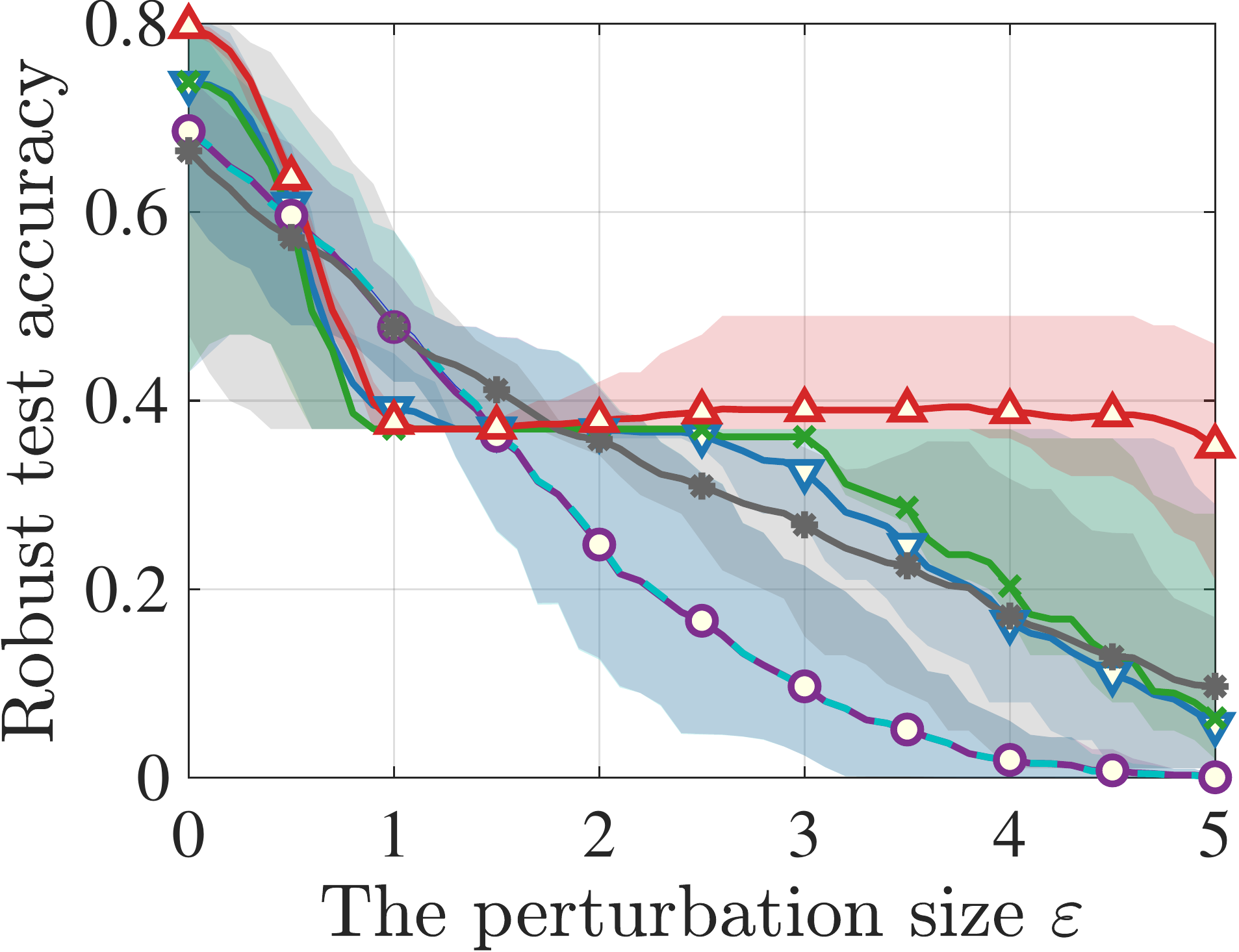}
\end{subfigure}
\caption{
Results on the synthetic dataset.
The first (resp., second) row reports the robust loss and accuracies on the training (resp., test) dataset. 
Each curve corresponds to the mean value over $6$ runs with the shaded area covering the minimum/maximum values.
}
\label{figure_toy}
\end{figure}

In this section, we conduct numerical experiments on synthetic and real datasets to demonstrate the efficiency of the proposed algorithm.\footnote{Source code: \url{https://github.com/xjiajiahao/fully-randomized-adversarial-training}}
We compare the proposed {\AlgAbbr} algorithm with SAT~\cite{madry2018towards}, Boosted Adversarial Training (BAT)~\cite{pinot2020randomization}, and the algorithms in~\cite{meunier2021mixed}.
Note that SAT produces a single deterministic classification model, BAT produces a mixture of $2$ models, and {\AlgAbbr} and those in~\cite{meunier2021mixed} produce mixtures of classification models with tunable sizes.

\subsection{Experiment on Synthetic Data}
In the first experiment, we consider training adversarially robust classifiers on a synthetic dataset following~\cite{meunier2021mixed}.
The synthetic data distribution $P(\xB, y)$ is constructed as follows. 
First, we let the feature space $\XM$ be $\RBB^2$ and the label space $\YM = \{-1, +1\}$.
Then, we set the marginal distribution of the label as $P(y = +1) = P(y = -1) = 1/2$ and set the conditional distribution of the features as $P(\xB | y = +1) = \frac{3}{4} \NM((3, 0), \IB_2) + \frac{1}{4} \NM((-3, 0), \IB_2)$ and $P(\xB | y = -1) = \NM((0, 0), \IB_2)$.
Here, $\IB_2$ denotes the identity matrix in $\RBB^{2 \times 2}$ and $\NM(\xB, \IB_2)$ denotes the Gaussian distribution with mean $\xB$ and covariance matrix $\IB_2$.
We generate the training dataset by randomly drawing $N = 100$ data samples from the distribution $P(\xB, y)$.
The test dataset is independently generated in the same way.
We use linear classification models of the form $\theta = (\wB^T, b)^T \in \Theta = \RBB^3$, where $\wB \in \RBB^2$ and $b \in \RBB$ are the weight and bias parameters, respectively. 
The loss function $\ell(\theta, (\xB, y))$ and the constraint set $\BBB_{\varepsilon}$ in~\eqref{eq_finite_problem} are defined as the logistic loss function and the $\ell_{2}$ norm ball with radius $\varepsilon$, respectively.

We compare {\AlgAbbr} with five baselines: SAT, BAT, and the oracle algorithm, the regularized algorithm, and the Adversarial Training of Mixtures (ATM) algorithm in~\cite{meunier2021mixed}.
Note that the oracle algorithm and the regularized algorithm are restricted to the case where $\Theta$ is finite discrete space.
To apply these two algorithms in our experiment, we follow~\cite{meunier2021mixed} and generate a finite discrete model space by randomly sampling $20$ linear models from $[-7, 7]^2$ with accuracies higher than $0.6$ on the clean training data.
For a fair comparison, we set the size of the mixture of models in {\AlgAbbr} and ATM to $20$ and initialize the $20$ models randomly.
In the implementation of {\AlgAbbr}, we use the PLA algorithm described previously as the sampling subroutine, and the expectation in~\eqref{eq_projectd_langevin_update} is estimated by drawing $100$ models from $\{\mu^{(0)}, \ldots, \mu^{(t)}\}$ when $t \ge 100$.
The regularization parameter in both {\AlgAbbr} and the regularized algorithm in~\cite{meunier2021mixed} is set to $0.01$.
In addition, to approximate the inner maximization problem in~\eqref{eq_finite_problem}, which is required by SAT, BAT, the oracle algorithm, and ATM, we first uniformly sample $1000$ points from $\BBB_{\varepsilon}(\xB_i)$ and then selects the one maximizing the loss, where $\xB_i$ is the feature vector of a sample to be attacked.
We run each algorithm until convergence under different perturbation sizes from the range $\{0, 0.1, 0.2, \ldots, 5.0\}$.

The experimental results are shown in Figure~\ref{figure_toy}.
Generally, it can be observed that {\AlgAbbr} achieves the lowest robust training/test loss and highest training/test accuracies among the $6$ algorithms, where the robust training/test loss is the loss value of the classifier on the perturbed training/test data samples generated by the adversary described above, and the robust training/test accuracy is defined accordingly.
Moreover, our {\AlgAbbr} algorithm constantly outperforms other methods by a large margin in terms of the robust training/test accuracies for large perturbation level ($\varepsilon > 2.5$).
We also observe that BAT slightly outperforms SAT in terms of the robust training and test accuracies, and ATM is slightly inferior to SAT under a large range of $\varepsilon$.
The oracle and the regularized algorithms, which only update the mixing weights of the initialized models and keep the model parameters fixed during training, are significantly inferior to other algorithms.
These two algorithms also have high variance among different runs, implying that their performance is sensitive to the quality of the initialized models.

\begin{figure}[t]
\begin{subfigure}{.49\linewidth}
  \centering
  \includegraphics[width=\linewidth]{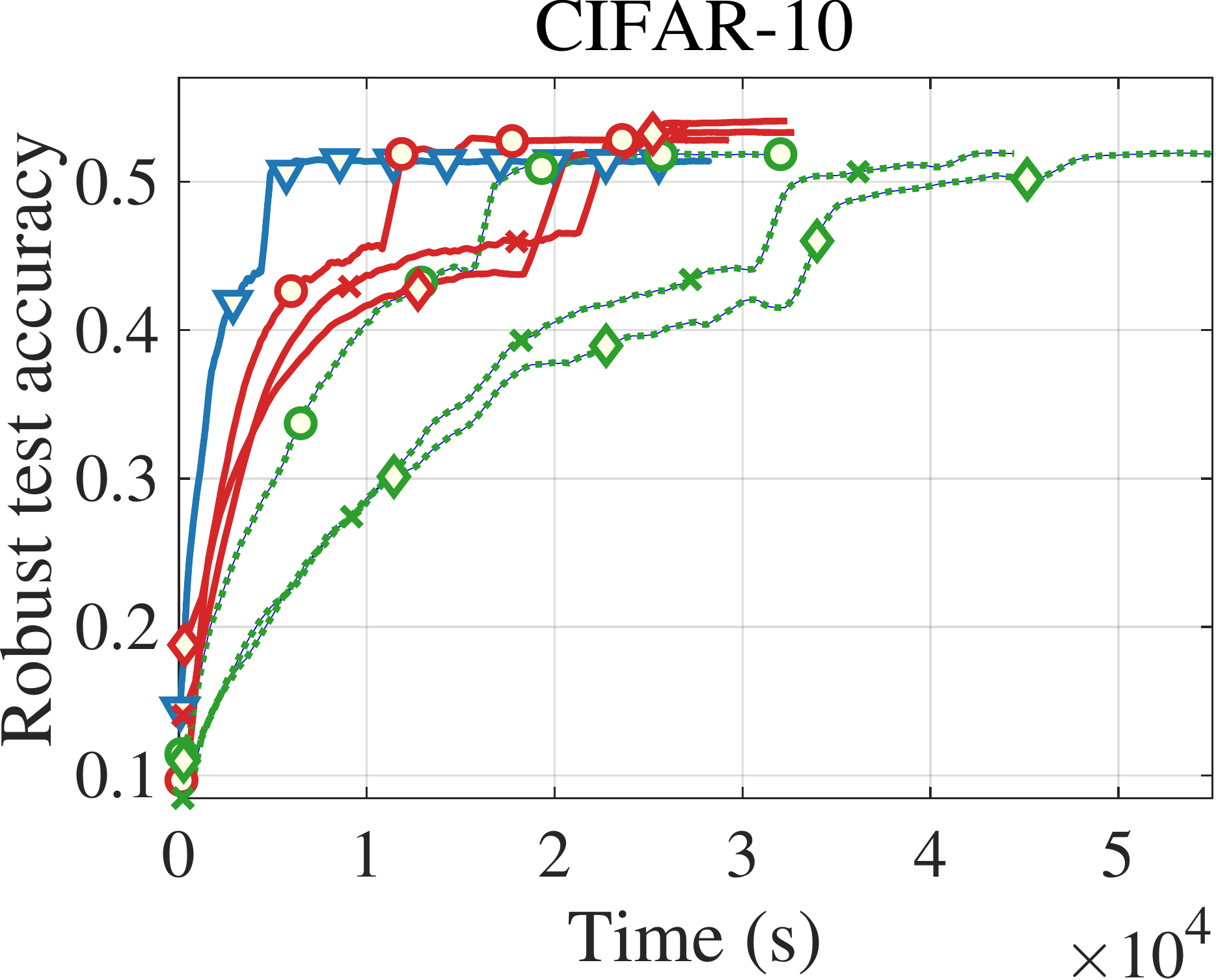}
\end{subfigure}
\hfill
\begin{subfigure}{.49\linewidth}
  \centering
  \includegraphics[width=\linewidth]{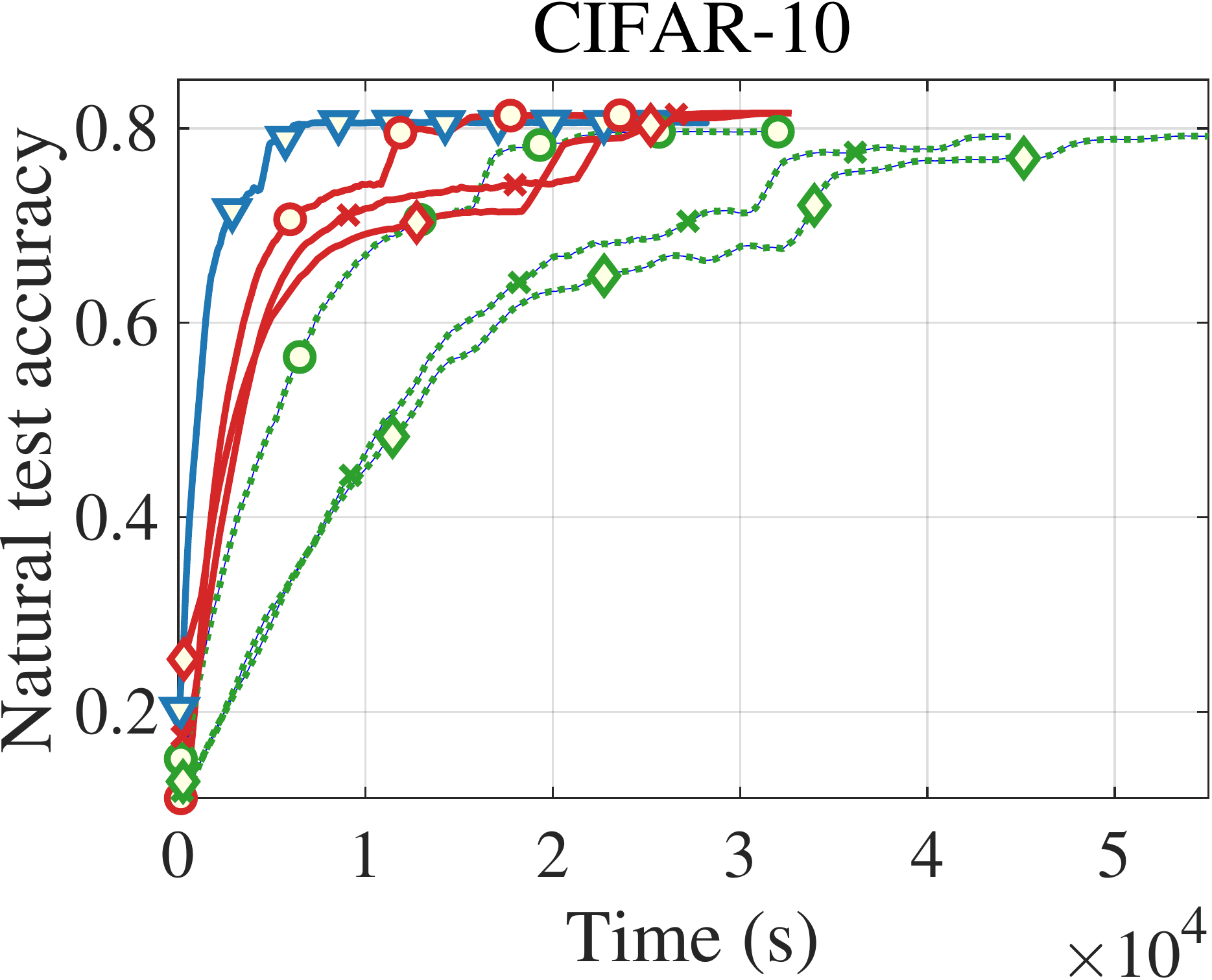}
\end{subfigure}
\\
\begin{subfigure}{.49\linewidth}
  \centering
  \includegraphics[width=\linewidth]{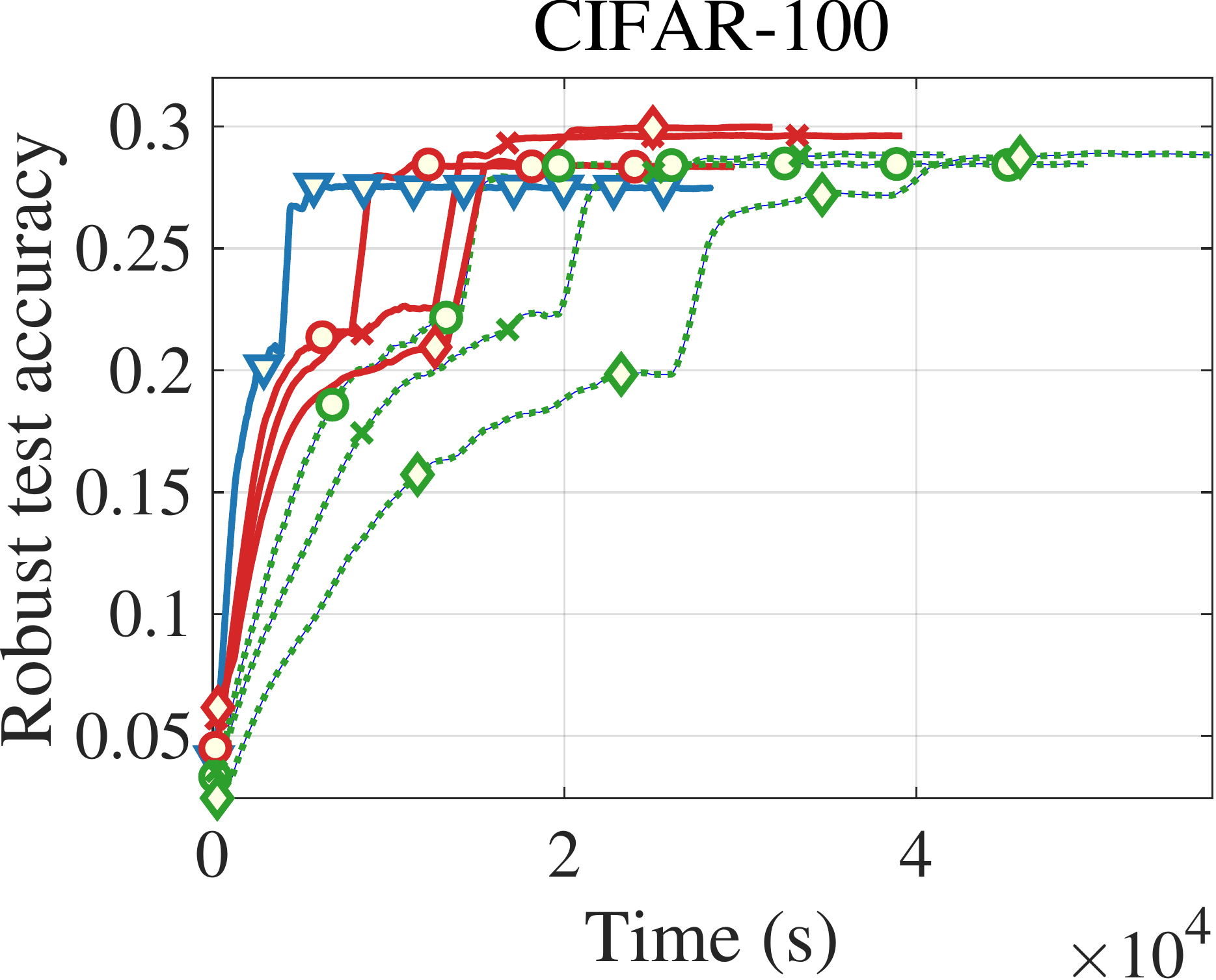}
\end{subfigure}
\hfill
\begin{subfigure}{.49\linewidth}
  \centering
  \includegraphics[width=\linewidth]{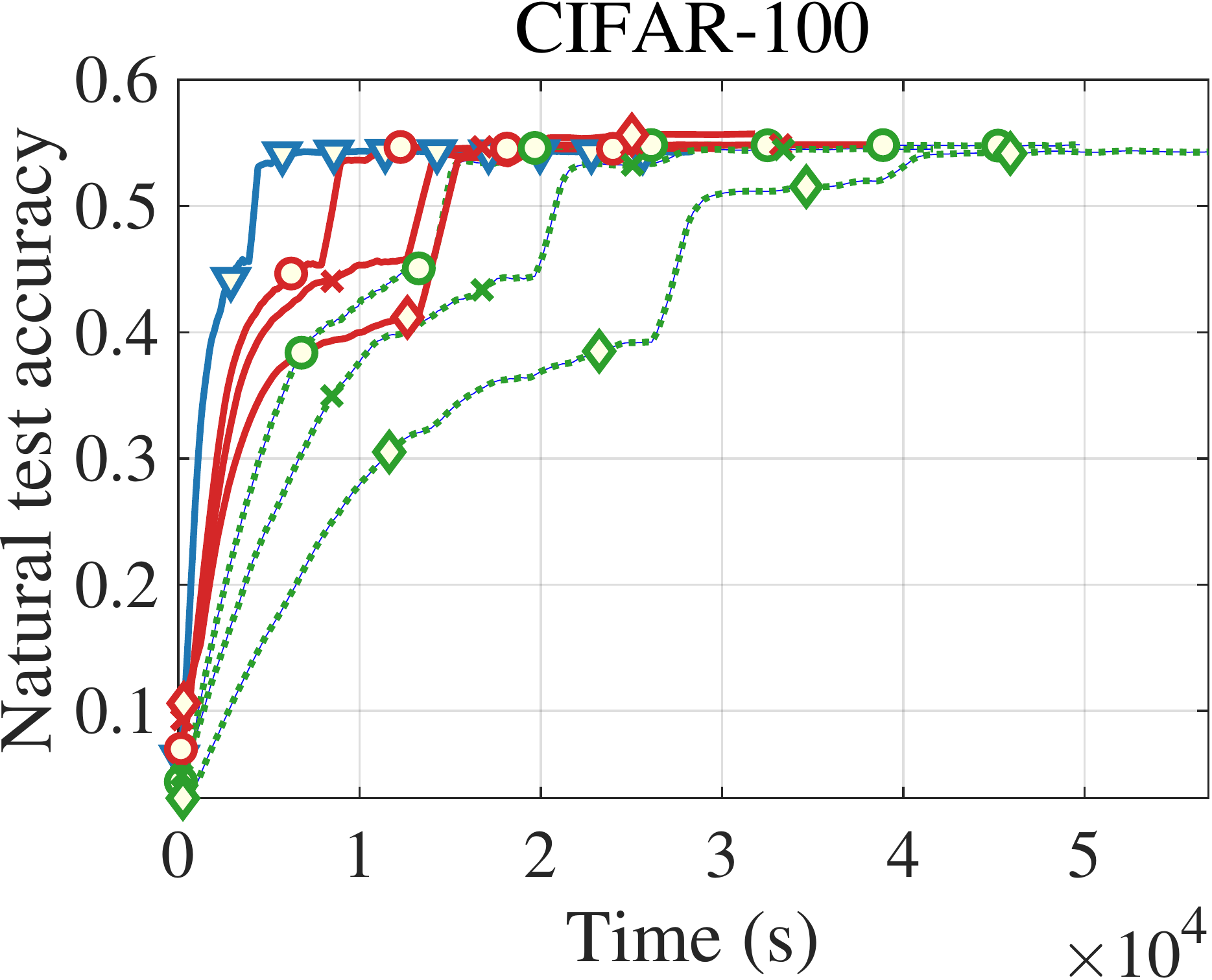}
\end{subfigure}
\\[0.2em]
\begin{subfigure}{\linewidth}
  \centering
  \includegraphics[trim={0cm 0cm 0cm 0cm}, clip, width=.98\linewidth]{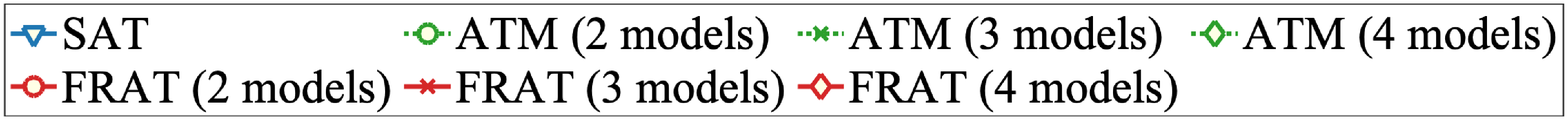}
\end{subfigure}
\caption{
Results on the CIFAR-10 (top) and CIFAR-100 (bottom) datasets.
The left (resp., right) column shows the test accuracy against $20$ steps of PGD attack (resp, the natural accuracy on the clean test data).
}
\label{figure_cifar}
\end{figure}

\begin{table*}[t]
    \centering
    \begin{tabular}{c|c|c|ccc}
        \toprule
        Dataset & Algorithm & Natural Test Accuracy & APGD\textsubscript{CE} & APGD\textsubscript{DLR} & APGD\textsubscript{CE} \& APGD\textsubscript{DLR} \\
        \midrule
        \multirow{4}{*}{CIFAR-10} & SAT & 80.6 & 49.7 & 48.1 & 47.6  \\
        & BAT & 80.6 & \bftab{53.3} & 45.2 & 45.0  \\
        & ATM ($2$ models) & 79.6 & 50.2 & 47.7 & 47.3  \\
        & {\AlgAbbr} ($2$ models)   & 81.1 & 50.9         & 49.3      & 48.5  \\
        & ATM ($3$ models) & 79.2 & 50.4 & 47.8 & 47.3  \\
        & {\AlgAbbr} ($3$ models)   & 81.1 & 51.5         & 49.7      & 48.9  \\
        & ATM ($4$ models) & 79.2 & 50.4 & 47.7 & 47.3  \\
        & {\AlgAbbr} ($4$ models)   & \bftab{81.6} & 52.1         & \bftab{50.1}      & \bftab{49.4}  \\
        \midrule
        \multirow{4}{*}{CIFAR-100} & SAT & 54.1 & 26.5 & 23.4 & 23.2  \\
        & BAT & 52.5 & \bftab{28.9} & 22.6 & 22.5  \\
        & ATM ($2$ models) & 54.9 & 27.5 & 24.1 & 23.7  \\
        & {\AlgAbbr} ($2$ models)  & 54.6 & 27.5 & 24.3 & 23.9  \\
        & ATM ($3$ models) & 54.5 & 27.9 & 24.4 & 24.1  \\
        & {\AlgAbbr} ($3$ models)   & 54.8 & 28.4 & 25.5 & 24.9  \\
        & ATM ($4$ models) & 54.2 & 27.9 & 24.6 & 24.2  \\
        & {\AlgAbbr} ($4$ models)   & \bftab{55.6} & 28.7 & \bftab{26.1} & \bftab{25.5}  \\
        \bottomrule
    \end{tabular}
    \caption{Results of the compared algorithms on CIFAR-10 and CIFAR-100 datasets.
    The third column shows the natural accuracy on the clean test data.
    The fourth (resp., fifth) column correspond to the robust test accuracies against AutoPGD\textsubscript{CE} (resp., APGD\textsubscript{DLR}).
    The last column presents the robust accuracy against the combination of AutoPGD\textsubscript{CE} and APGD\textsubscript{DLR} attacks.
    }
    \label{table_cifar}
\end{table*}

\subsection{Experiments on Real Data}
To demonstrate the efficiency of the proposed algorithm in practice, we conduct experiments on CIFAR-10 and CIFAR-100 datasets~\cite{krizhevsky2009learning}.
We compare {\AlgAbbr} with SAT, BAT, and ATM.
The oracle and regularized algorithms in~\cite{meunier2021mixed} are excluded from our baselines as they are impractical in high-dimensional spaces.
For ATM and {\AlgAbbr}, we test different sizes $M$ of the mixture in the range $\{2, 3, 4\}$, and denote the corresponding algorithms as ATM/FRAT (M models).
We basically follow the experimental setting in~\cite{meunier2021mixed}.
The detailed setting is deferred to the Appendix~\ref{section_exp_setting}.
For {\AlgAbbr}, we implement the sampling subroutine with $10$ steps of PLA~\eqref{eq_projectd_langevin_update}, where the noise level $\gamma$ is set to $0.0001$ and the expectation over $\bar{\mu}^{(t)}$ is approximated with the sliding window trick described previously.
We set the size of the sliding window to $1$, which already performs well.
The average runtime per iteration of SAT is 0.72 s on CIFAR-10 (resp., 1.63 s on CIFAR-100); FRAT with M = 2, 3, and 4 models take 1.50 s, 2.24 s, and 3.19 s on CIFAR-10 (resp., 1.63 s, 2.30 s, and 2.87 s on CIFAR-100), which are about M times of SAT and corroborate our analysis.

After training, we evaluate the classifier obtained by each algorithm using an adapted version of AutoPGD untargeted attacks~\cite{croce2020reliable} with both Cross Entropy (CE) and Difference of Logits Ratio (DLR) loss.
We refer to these two attacks as APGD\textsubscript{CE} and APGD\textsubscript{DLR}.

The training curves of SAT, ATM, and {\AlgAbbr} are shown in Figure~\ref{figure_cifar}, where the robust test accuracy is the accuracy of a classifier on perturbed data samples generated by $20$ steps of PGD attack (PGD\textsubscript{$20$} for short), and the natural test accuracy is the accuracy on clean data samples.
Note that the result of BAT is excluded, because it is not an iterative algorithm but rather a one-step boosting method based on SAT.
In addition, we terminate the training when the performance plateaus.
We observe that both ATM and {\AlgAbbr} with mixture sizes in the range $\{2, 3, 4\}$ achieve higher robust test accuracies than that of SAT, while the natural test accuracies of all algorithms are similar.
This demonstrates the efficiency of using randomized strategies.
Moreover, {\AlgAbbr} converges faster than ATM in terms of both the robust and natural accuracies when they use the same mixture size, and the robust test accuracy of {\AlgAbbr} at convergence is higher than that of ATM.
Note that as the size $M$ of the mixture increases, the convergence rate of ATM slows down significantly, whereas the convergence rate of {\AlgAbbr} is less affected by the size $M$.

Table~\ref{table_cifar} compares the performance of SAT, BAT, ATM, and {\AlgAbbr} after training.
We can see that as $M$ increases, the performance of {\AlgAbbr} improves, and
{\AlgAbbr} with $M = 4$ achieves the best natural test accuracy, the accuracy against APGD\textsubscript{DLR}, and the accuracy against APGD\textsubscript{CE} \& APGD\textsubscript{DLR}.
We also observe that {\AlgAbbr} outperforms ATM when they use the same size $M$, and {\AlgAbbr} with $M = 2$ already outperforms SAT and BAT.
These results demonstrate the robustness of {\AlgAbbr} against strong attacks, and the superiority of {\AlgAbbr} over SAT, BAT, and ATM.
Note that as the training procedure of BAT uses the CE loss to generate adversarial examples, BAT achieves the best performance in terms of the accuracy against APGD\textsubscript{CE}.
However, BAT performs worse in terms of the accuracy against the APGD\textsubscript{DLR} attack and the APGD\textsubscript{CE} \& APGD\textsubscript{CE} attack, which indicates that BAT is vulnerable to general attacks other than CE.


\section{Conclusion}  \label{section_conclusion}

In this paper, we propose an efficient algorithm for finding optimal randomized strategies in the AT game.
Our algorithm {\AlgAbbr} is obtained by discretizing a new continuous-time and interacting flow that properly exploits the problem structure.
We prove that this flow converges to an MNE at a sublinear rate.
Experimental results demonstrate the advantages of {\AlgAbbr} over existing ones.

\clearpage

\section*{Acknowledgements}
This work is supported by National Key Research and Development Program of China under Grant 2020AAA0107400, Zhejiang Provincial Natural Science Foundation of China under Grant No. LZ18F020002, and National Natural Science Foundation of China (Grant No: 62206248).

\bibliography{references}

\begin{thebibliography}{46}
\providecommand{\natexlab}[1]{#1}

\bibitem[{Bai et~al.(2021)Bai, Luo, Zhao, Wen, and Wang}]{bai2021recent}
Bai, T.; Luo, J.; Zhao, J.; Wen, B.; and Wang, Q. 2021.
\newblock Recent Advances in Adversarial Training for Adversarial Robustness.
\newblock In \emph{Proceedings of the Thirtieth International Joint Conference on Artificial Intelligence, {IJCAI-21}}, 4312--4321.

\bibitem[{Biggio et~al.(2013)Biggio, Corona, Maiorca, Nelson, {\v{S}}rndi{\'c}, Laskov, Giacinto, and Roli}]{biggio2013evasion}
Biggio, B.; Corona, I.; Maiorca, D.; Nelson, B.; {\v{S}}rndi{\'c}, N.; Laskov, P.; Giacinto, G.; and Roli, F. 2013.
\newblock Evasion attacks against machine learning at test time.
\newblock In \emph{Joint European conference on machine learning and knowledge discovery in databases}, 387--402. Springer.

\bibitem[{Bonesky et~al.(2007)Bonesky, Bredies, Lorenz, and Maass}]{bonesky2007generalized}
Bonesky, T.; Bredies, K.; Lorenz, D.~A.; and Maass, P. 2007.
\newblock A generalized conditional gradient method for nonlinear operator equations with sparsity constraints.
\newblock \emph{Inverse Problems}, 23(5): 2041.

\bibitem[{Bose et~al.(2020)Bose, Gidel, Berard, Cianflone, Vincent, Lacoste-Julien, and Hamilton}]{bose2020adversarial}
Bose, J.; Gidel, G.; Berard, H.; Cianflone, A.; Vincent, P.; Lacoste-Julien, S.; and Hamilton, W. 2020.
\newblock Adversarial example games.
\newblock \emph{Advances in neural information processing systems}, 33: 8921--8934.

\bibitem[{Bredies, Lorenz, and Maass(2009)}]{bredies2009generalized}
Bredies, K.; Lorenz, D.~A.; and Maass, P. 2009.
\newblock A generalized conditional gradient method and its connection to an iterative shrinkage method.
\newblock \emph{Computational Optimization and applications}, 42(2): 173--193.

\bibitem[{Bubeck, Eldan, and Lehec(2018)}]{bubeck2018sampling}
Bubeck, S.; Eldan, R.; and Lehec, J. 2018.
\newblock Sampling from a log-concave distribution with projected langevin monte carlo.
\newblock \emph{Discrete \& Computational Geometry}, 59(4): 757--783.

\bibitem[{Bul{\`o} et~al.(2016)Bul{\`o}, Biggio, Pillai, Pelillo, and Roli}]{bulo2016randomized}
Bul{\`o}, S.~R.; Biggio, B.; Pillai, I.; Pelillo, M.; and Roli, F. 2016.
\newblock Randomized prediction games for adversarial machine learning.
\newblock \emph{IEEE transactions on neural networks and learning systems}, 28(11): 2466--2478.

\bibitem[{Chizat(2022)}]{chizat2022sparse}
Chizat, L. 2022.
\newblock Sparse optimization on measures with over-parameterized gradient descent.
\newblock \emph{Mathematical Programming}, 194(1): 487--532.

\bibitem[{Cohen, Rosenfeld, and Kolter(2019)}]{cohen2019certified}
Cohen, J.; Rosenfeld, E.; and Kolter, Z. 2019.
\newblock Certified adversarial robustness via randomized smoothing.
\newblock In \emph{International Conference on Machine Learning}, 1310--1320. PMLR.

\bibitem[{Croce and Hein(2020)}]{croce2020reliable}
Croce, F.; and Hein, M. 2020.
\newblock Reliable evaluation of adversarial robustness with an ensemble of diverse parameter-free attacks.
\newblock In \emph{International conference on machine learning}, 2206--2216. PMLR.

\bibitem[{Dhillon et~al.(2018)Dhillon, Azizzadenesheli, Lipton, Bernstein, Kossaifi, Khanna, and Anandkumar}]{dhillon2018stochastic}
Dhillon, G.~S.; Azizzadenesheli, K.; Lipton, Z.~C.; Bernstein, J.~D.; Kossaifi, J.; Khanna, A.; and Anandkumar, A. 2018.
\newblock Stochastic Activation Pruning for Robust Adversarial Defense.
\newblock In \emph{International Conference on Learning Representations}.

\bibitem[{Domingo-Enrich et~al.(2020)Domingo-Enrich, Jelassi, Mensch, Rotskoff, and Bruna}]{domingo2020mean}
Domingo-Enrich, C.; Jelassi, S.; Mensch, A.; Rotskoff, G.; and Bruna, J. 2020.
\newblock A mean-field analysis of two-player zero-sum games.
\newblock \emph{Advances in neural information processing systems}, 33: 20215--20226.

\bibitem[{Goodfellow, Shlens, and Szegedy(2015)}]{goodfellow2015explaining}
Goodfellow, I.~J.; Shlens, J.; and Szegedy, C. 2015.
\newblock Explaining and harnessing adversarial examples.
\newblock In \emph{International Conference on Learning Representations}.

\bibitem[{Gowal et~al.(2021)Gowal, Rebuffi, Wiles, Stimberg, Calian, and Mann}]{gowal2021improving}
Gowal, S.; Rebuffi, S.-A.; Wiles, O.; Stimberg, F.; Calian, D.~A.; and Mann, T.~A. 2021.
\newblock Improving robustness using generated data.
\newblock \emph{Advances in Neural Information Processing Systems}, 34: 4218--4233.

\bibitem[{He et~al.(2016)He, Zhang, Ren, and Sun}]{he2016deep}
He, K.; Zhang, X.; Ren, S.; and Sun, J. 2016.
\newblock Deep residual learning for image recognition.
\newblock In \emph{Proceedings of the IEEE conference on computer vision and pattern recognition}, 770--778.

\bibitem[{Hofbauer and Sandholm(2002)}]{hofbauer2002global}
Hofbauer, J.; and Sandholm, W.~H. 2002.
\newblock On the global convergence of stochastic fictitious play.
\newblock \emph{Econometrica}, 70(6): 2265--2294.

\bibitem[{Hsieh, Liu, and Cevher(2019)}]{hsieh2019finding}
Hsieh, Y.-P.; Liu, C.; and Cevher, V. 2019.
\newblock Finding mixed nash equilibria of generative adversarial networks.
\newblock In \emph{International Conference on Machine Learning}, 2810--2819. PMLR.

\bibitem[{Krizhevsky, Hinton et~al.(2009)}]{krizhevsky2009learning}
Krizhevsky, A.; Hinton, G.; et~al. 2009.
\newblock Learning multiple layers of features from tiny images.
\newblock Citeseer.

\bibitem[{Lahkar and Riedel(2014)}]{lahkar2014continuous}
Lahkar, R.; and Riedel, F. 2014.
\newblock The continuous logit dynamic and price dispersion.
\newblock \emph{Institute of Mathematical Economics Working Paper}, (521).

\bibitem[{Lahkar and Riedel(2015)}]{lahkar2015logit}
Lahkar, R.; and Riedel, F. 2015.
\newblock The logit dynamic for games with continuous strategy sets.
\newblock \emph{Games and Economic Behavior}, 91: 268--282.

\bibitem[{Liero, Mielke, and Savar{\'e}(2018)}]{liero2018optimal}
Liero, M.; Mielke, A.; and Savar{\'e}, G. 2018.
\newblock Optimal entropy-transport problems and a new Hellinger--Kantorovich distance between positive measures.
\newblock \emph{Inventiones mathematicae}, 211(3): 969--1117.

\bibitem[{Liu et~al.(2021)Liu, Zhang, Yang, Babanezhad, and Wang}]{liu2021infinite}
Liu, L.; Zhang, Y.; Yang, Z.; Babanezhad, R.; and Wang, Z. 2021.
\newblock Infinite-Dimensional Optimization for Zero-Sum Games via Variational Transport.
\newblock In \emph{International Conference on Machine Learning}, 7033--7044. PMLR.

\bibitem[{Liu(2017)}]{liu2017stein}
Liu, Q. 2017.
\newblock Stein variational gradient descent as gradient flow.
\newblock \emph{Advances in neural information processing systems}, 30.

\bibitem[{Liutkus et~al.(2019)Liutkus, Simsekli, Majewski, Durmus, and St{\"o}ter}]{liutkus2019sliced}
Liutkus, A.; Simsekli, U.; Majewski, S.; Durmus, A.; and St{\"o}ter, F.-R. 2019.
\newblock Sliced-Wasserstein flows: Nonparametric generative modeling via optimal transport and diffusions.
\newblock In \emph{International Conference on Machine Learning}, 4104--4113. PMLR.

\bibitem[{Ma and Ying(2021)}]{ma2021provably}
Ma, C.; and Ying, L. 2021.
\newblock Provably convergent quasistatic dynamics for mean-field two-player zero-sum games.
\newblock In \emph{International Conference on Learning Representations}.

\bibitem[{Madry et~al.(2018)Madry, Makelov, Schmidt, Tsipras, and Vladu}]{madry2018towards}
Madry, A.; Makelov, A.; Schmidt, L.; Tsipras, D.; and Vladu, A. 2018.
\newblock Towards Deep Learning Models Resistant to Adversarial Attacks.
\newblock In \emph{International Conference on Learning Representations}.

\bibitem[{Maini, Wong, and Kolter(2020)}]{maini2020adversarial}
Maini, P.; Wong, E.; and Kolter, Z. 2020.
\newblock Adversarial robustness against the union of multiple perturbation models.
\newblock In \emph{International Conference on Machine Learning}, 6640--6650. PMLR.

\bibitem[{Meunier et~al.(2021)Meunier, Scetbon, Pinot, Atif, and Chevaleyre}]{meunier2021mixed}
Meunier, L.; Scetbon, M.; Pinot, R.~B.; Atif, J.; and Chevaleyre, Y. 2021.
\newblock Mixed nash equilibria in the adversarial examples game.
\newblock In \emph{International Conference on Machine Learning}, 7677--7687. PMLR.

\bibitem[{Moosavi-Dezfooli et~al.(2019)Moosavi-Dezfooli, Fawzi, Uesato, and Frossard}]{moosavi2019robustness}
Moosavi-Dezfooli, S.-M.; Fawzi, A.; Uesato, J.; and Frossard, P. 2019.
\newblock Robustness via curvature regularization, and vice versa.
\newblock In \emph{Proceedings of the IEEE/CVF Conference on Computer Vision and Pattern Recognition}, 9078--9086.

\bibitem[{Pang et~al.(2020)Pang, Yang, Dong, Su, and Zhu}]{pang2020bag}
Pang, T.; Yang, X.; Dong, Y.; Su, H.; and Zhu, J. 2020.
\newblock Bag of Tricks for Adversarial Training.
\newblock In \emph{International Conference on Learning Representations}.

\bibitem[{Papernot et~al.(2016)Papernot, McDaniel, Wu, Jha, and Swami}]{papernot2016distillation}
Papernot, N.; McDaniel, P.; Wu, X.; Jha, S.; and Swami, A. 2016.
\newblock Distillation as a defense to adversarial perturbations against deep neural networks.
\newblock In \emph{2016 IEEE symposium on security and privacy (SP)}, 582--597. IEEE.

\bibitem[{Perdomo and Singer(2019)}]{perdomo2019robust}
Perdomo, J.~C.; and Singer, Y. 2019.
\newblock Robust attacks against multiple classifiers.
\newblock \emph{arXiv preprint arXiv:1906.02816}.

\bibitem[{Perkins and Leslie(2014)}]{perkins2014stochastic}
Perkins, S.; and Leslie, D.~S. 2014.
\newblock Stochastic fictitious play with continuous action sets.
\newblock \emph{Journal of Economic Theory}, 152: 179--213.

\bibitem[{Pinot et~al.(2020)Pinot, Ettedgui, Rizk, Chevaleyre, and Atif}]{pinot2020randomization}
Pinot, R.; Ettedgui, R.; Rizk, G.; Chevaleyre, Y.; and Atif, J. 2020.
\newblock Randomization matters how to defend against strong adversarial attacks.
\newblock In \emph{International Conference on Machine Learning}, 7717--7727. PMLR.

\bibitem[{Pinot et~al.(2019)Pinot, Meunier, Araujo, Kashima, Yger, Gouy-Pailler, and Atif}]{pinot2019theoretical}
Pinot, R.; Meunier, L.; Araujo, A.; Kashima, H.; Yger, F.; Gouy-Pailler, C.; and Atif, J. 2019.
\newblock Theoretical evidence for adversarial robustness through randomization.
\newblock \emph{Advances in Neural Information Processing Systems}, 32.

\bibitem[{Rotskoff et~al.(2019)Rotskoff, Jelassi, Bruna, and Vanden-Eijnden}]{rotskoff2019global}
Rotskoff, G.; Jelassi, S.; Bruna, J.; and Vanden-Eijnden, E. 2019.
\newblock Global convergence of neuron birth-death dynamics.
\newblock In \emph{International Conference on Machine Learning}.

\bibitem[{Samangouei, Kabkab, and Chellappa(2018)}]{samangouei2018defense}
Samangouei, P.; Kabkab, M.; and Chellappa, R. 2018.
\newblock Defense-GAN: Protecting Classifiers Against Adversarial Attacks Using Generative Models.
\newblock In \emph{International Conference on Learning Representations}.

\bibitem[{Schott et~al.(2018)Schott, Rauber, Bethge, and Brendel}]{schott2018towards}
Schott, L.; Rauber, J.; Bethge, M.; and Brendel, W. 2018.
\newblock Towards the first adversarially robust neural network model on MNIST.
\newblock In \emph{International Conference on Learning Representations}.

\bibitem[{Shaham, Yamada, and Negahban(2018)}]{shaham2018understanding}
Shaham, U.; Yamada, Y.; and Negahban, S. 2018.
\newblock Understanding adversarial training: Increasing local stability of supervised models through robust optimization.
\newblock \emph{Neurocomputing}, 307: 195--204.

\bibitem[{Su, Boyd, and Candes(2014)}]{su2014differential}
Su, W.; Boyd, S.; and Candes, E. 2014.
\newblock A differential equation for modeling Nesterov’s accelerated gradient method: theory and insights.
\newblock \emph{Advances in neural information processing systems}, 27.

\bibitem[{Suggala and Netrapalli(2020)}]{suggala2020follow}
Suggala, A.; and Netrapalli, P. 2020.
\newblock Follow the perturbed leader: Optimism and fast parallel algorithms for smooth minimax games.
\newblock \emph{Advances in Neural Information Processing Systems}, 33: 22316--22326.

\bibitem[{Szegedy et~al.(2014)Szegedy, Zaremba, Sutskever, Bruna, Erhan, Goodfellow, and Fergus}]{szegedy2014intriguing}
Szegedy, C.; Zaremba, W.; Sutskever, I.; Bruna, J.; Erhan, D.; Goodfellow, I.; and Fergus, R. 2014.
\newblock Intriguing properties of neural networks.
\newblock In \emph{International Conference on Learning Representations}.

\bibitem[{Wang, Shi, and Osher(2019)}]{wang2019resnets}
Wang, B.; Shi, Z.; and Osher, S. 2019.
\newblock Resnets ensemble via the feynman-kac formalism to improve natural and robust accuracies.
\newblock \emph{Advances in Neural Information Processing Systems}, 32.

\bibitem[{Welling and Teh(2011)}]{welling2011bayesian}
Welling, M.; and Teh, Y.~W. 2011.
\newblock Bayesian learning via stochastic gradient Langevin dynamics.
\newblock In \emph{Proceedings of the 28th international conference on machine learning (ICML-11)}, 681--688. Citeseer.

\bibitem[{Xie et~al.(2018)Xie, Wang, Zhang, Ren, and Yuille}]{xie2018mitigating}
Xie, C.; Wang, J.; Zhang, Z.; Ren, Z.; and Yuille, A. 2018.
\newblock Mitigating Adversarial Effects Through Randomization.
\newblock In \emph{International Conference on Learning Representations}.

\bibitem[{Zhang et~al.(2019)Zhang, Yu, Jiao, Xing, El~Ghaoui, and Jordan}]{zhang2019theoretically}
Zhang, H.; Yu, Y.; Jiao, J.; Xing, E.; El~Ghaoui, L.; and Jordan, M. 2019.
\newblock Theoretically principled trade-off between robustness and accuracy.
\newblock In \emph{International conference on machine learning}, 7472--7482. PMLR.

\end{thebibliography}

\clearpage
\onecolumn
\appendix

\section{The Existence of Mixed Nash Equilibria}  \label{section_existence_of_mne}
\citet{meunier2021mixed} prove the existence of mixed Nash equilibria in a randomized AT game whose problem formulation is slightly different from our formulation~\eqref{eq_infinite_problem}.
Nevertheless, the existence of MNE in~\eqref{eq_infinite_problem} directly follows from the result of~\cite{meunier2021mixed} as we show in what follows.
First, we restate the result of~\cite{meunier2021mixed} in the following proposition, which is obtained by their Theorem 1 and Proposition 8.

\begin{proposition}  \label{proposition_existence_of_mne} 
    Suppose that $\ell: \Theta \times (\XM \times \YM) \rightarrow \RBB$ satisfies the following conditions: (i) the function $\ell$ is non-negative and Borel measurable, (ii) $\ell(\theta, \cdot)$ is upper-semi continuous for any $\theta \in \Theta$, and (iii) $\exists U > 0$, $\forall \theta \in \Theta$ and $(\xB, y) \in \XM \times \YM$, $0 \le \ell(\theta, (\xB, y)) \le U$.
    Define $\Gamma_{\varepsilon} = (\Gamma_{1, \varepsilon}, \ldots, \Gamma_{N, \varepsilon})$ with $\Gamma_{i, \varepsilon} = \{\QBB_i \in \MM_1^+(\XM \times \YM): \int c_{\varepsilon}((\xB_i, y_i), \cdot) = 0\}$,
    where $c_{\varepsilon}((\xB_i, y_i), \cdot)$ is given by 
    \begin{align*}
        c_{\varepsilon}((\xB_i, y_i), (\xB', y')) = 
        \begin{cases}
            0, \quad \; \textrm{if} \ d(\xB, \xB') \le \varepsilon \ \textrm{and} \ y = y' \\
            +\infty, \ \textrm{otherwise}.
        \end{cases}
    \end{align*}
    Then the following strong duality holds
    \begin{align*}
        \inf_{\mu \in \MM_1^+(\Theta)} \sup_{\QBB \in \Gamma_{\varepsilon}} \frac{1}{N} \sum_{i=1}^N \EBB_{(\xB, y) \sim \QBB_i, \theta \sim \mu} [ \ell(\theta, (\xB, y))]
        = \sup_{\QBB \in \Gamma_{\varepsilon}} \inf_{\mu \in \MM_1^+(\Theta)} \frac{1}{N} \sum_{i=1}^N \EBB_{(\xB, y) \sim \QBB_i, \theta \sim \mu} [ \ell(\theta, (\xB, y))].
    \end{align*}
    where the supremum in~\eqref{eq_infinite_problem} is always attained.
    Further, if $\Theta$ is a compact set, and $\forall (\xB, y) \in \XM \times \YM$, $\ell(\cdot, (\xB, y))$ is lower semi-continuous, the infimum is also attained.
\end{proposition}




Now, we show that there also exists an MNE in the zero-sum game ~\eqref{eq_infinite_problem}.
Note that the conditions in Proposition~\ref{proposition_existence_of_mne} are indeed weaker than those in Assumption~\ref{assumption_loss_function}-\ref{assumption_initial_value}.
In addition, for any $\QBB_i \in \Gamma_{i, \varepsilon}$, $\QBB_i (\BBB_{\varepsilon}(\xB_i) \times \{y_i\}) = 1$,
which implies
\begin{align*}
    \sup_{\QBB_i \in \Gamma_{i, \varepsilon}} \EBB_{(\xB, y) \sim \QBB_i} [\ell(\theta, (\xB, y))]
    = \sup_{\nu_i \in \MM_1^+(\BBB_{\varepsilon}(\xB_i))} \EBB_{\xB \sim \nu_i}[ \ell(\theta, (\xB, y_i))].   
\end{align*}
On the other hand, for any $\QBB_i \in \Gamma_{i, \varepsilon}$, there exists a $\nu_i \in \MM_1^+(\BBB_{\varepsilon}(\xB_i))$ such that $\EBB_{(\xB, y) \sim \QBB_i} [\ell(\theta, (\xB, y))] = \EBB_{\xB \sim \nu_i}[ \ell(\theta, (\xB, y_i))]$.
Therefore, we have
\begin{align*}
    \sup_{\QBB \in \Gamma_{\varepsilon}} \inf_{\mu \in \MM_1^+(\Theta)} \frac{1}{N} \sum_{i=1}^N \EBB_{(\xB, y) \sim \QBB_i, \theta \sim \mu} [ \ell(\theta, (\xB, y))]
    \le \sup_{\nu \in \Sigma} \inf_{\mu \in \MM_1^+(\Theta)} \frac{1}{N} \sum_{i=1}^N \EBB_{\xB \sim \nu_i, \theta \sim \mu} [ \ell(\theta, (\xB, y))].
\end{align*}
Combining Proposition~\ref{proposition_existence_of_mne}, the above two formulae, and the fact that 
$
    \sup_{\nu \in \Sigma} \inf_{\mu \in \MM_1^+(\Theta)} \LM(\mu, \nu)
    \le \inf_{\mu \in \MM_1^+(\Theta)} \sup_{\nu \in \Sigma} \LM(\mu, \nu)
$,
we arrive at
\begin{align*}
    \inf_{\mu \in \MM_1^+(\Theta)} \sup_{\nu \in \Sigma} \LM(\mu, \nu)
    = \sup_{\nu \in \Sigma} \inf_{\mu \in \MM_1^+(\Theta)} \LM(\mu, \nu).
\end{align*}
Thus, there exists an MNE in~\eqref{eq_infinite_problem}.
Following the above analysis, one can also prove the existence of MNE in the regularized game~\eqref{eq_regularized_problem}.

\section{Deferred Proofs}  \label{proofs}

\subsection{Proof of Theorem~\ref{theorem_regularization_error}}
\begin{proof}
To begin with, we bound $\GM_{0}(\tilde{\mu}, \tilde{\nu})$ from above by
\begin{align}  \label{eq_regularization_error_init}
    &\GM_{0}(\tilde{\mu}, \tilde{\nu})  \nonumber \\
    ={}& \max_{\nu \in \Sigma} \LM(\tilde{\mu}, \nu) - \inf_{\mu} \LM(\mu, \tilde{\nu})  \nonumber \\
    ={}& \max_{\nu \in \Sigma} \LM(\tilde{\mu}, \nu)
        + \max_{\nu} \{\LM(\tilde{\mu}, \nu) - \beta \HM(\nu)\}
        - \max_{\nu} \{\LM(\tilde{\mu}, \nu) - \beta \HM(\nu)\}
        - \inf_{\mu} \LM(\mu, \tilde{\nu})  \nonumber \\
    \le{}& \max_{\nu \in \Sigma} \LM(\tilde{\mu}, \nu)
        - \max_{\nu} \{\LM(\tilde{\mu}, \nu) - \beta \HM(\nu)\}
        + \max_{\nu} \{\LM(\tilde{\mu}, \nu) - \beta \HM(\nu)\}
        - \inf_{\mu} \LM(\mu, \tilde{\nu})
        + \beta \HM(\tilde{\nu})  \nonumber \\
    \le{}& \max_{\nu \in \Sigma} \LM(\tilde{\mu}, \nu)
        - \max_{\nu} \{\LM(\tilde{\mu}, \nu) - \beta \HM(\nu)\}
        + \GM_{\beta}(\tilde{\mu}, \tilde{\nu}),  \nonumber \\
    ={}& \frac{1}{N} \sum_{i=1}^N \left(
        \max_{\nu_i \in \MM_1^+(\BBB_{\varepsilon}(\xB_i))} \{\EBB_{\theta \sim \tilde{\mu}, \xB \sim \nu_i} [\ell(\theta, (\xB, y_i))]\}
        - \max_{\nu_i \in \MM_1^+(\BBB_{\varepsilon}(\xB_i))} \left\{\EBB_{\theta \sim \tilde{\mu}, \xB \sim \nu_i} [\ell(\theta, (\xB, y_i))] - \beta \textrm{KL}(\nu_i \| u_i) \right\}
        \right)
        + \GM_{\beta}(\tilde{\mu}, \tilde{\nu}),
\end{align}
where the first inequality follows from the fact that $\HM(\nu) = \frac{1}{N} \sum_{i=1}^N \textrm{KL}(\nu_i \| u_i) \ge 0$.
Let $q_i(\xB)$ be the density function of $\nu_i$ at $\xB$.
Notice that $\max_{\nu_i} \left\{\EBB_{\theta \sim \tilde{\mu}, \xB \sim \nu_i} [\ell(\theta, (\xB, y_i))] - \beta \textrm{KL}(\nu_i \| u_i) \right\}$
has the following closed form
\begin{align}  \label{eq_sup_problem_closed_form}
    &\max_{\nu_i} \left\{\EBB_{\theta \sim \tilde{\mu}, \xB \sim \nu_i} [\ell(\theta, (\xB, y_i))] - \beta \textrm{KL}(\nu_i \| u_i) \right\}  \nonumber \\
    ={}& \max_{\nu_i} \Big\{ - \beta \int_{\BBB_{\varepsilon}(\xB_i)} q_i(\xB) \log q_i(\xB) d \xB
        + \beta \int_{\BBB_{\varepsilon}(\xB_i)} q_i(\xB) \log
            \frac{\exp(\beta^{-1} \EBB_{\theta \sim \tilde{\mu}}[\ell(\theta, (\xB, y_i))])}{\int_{\BBB_{\varepsilon}(\xB_i)} \exp(\beta^{-1} \EBB_{\theta \sim \tilde{\mu}}[\ell(\theta, (\xB, y_i))]) d \xB}
        d \xB  \nonumber \\
        &- \beta \log \textrm{Vol}(\BBB_{\varepsilon}(\xB_i))
        + \beta \log \Big( \int_{\BBB_{\varepsilon}(\xB_i)} \exp(\beta^{-1} \EBB_{\theta \sim \tilde{\mu}}[\ell(\theta, (\xB, y_i))]) d \xB \Big)
    \Big\}  \nonumber \\
    ={}& \beta \log \Big( \frac{\int_{\BBB_{\varepsilon}(\xB_i)} \exp(\beta^{-1} \EBB_{\theta \sim \tilde{\mu}}[\ell(\theta, (\xB, y_i))]) d \xB}{\textrm{Vol}(\BBB_{\varepsilon}(\xB_i))}\Big).
\end{align}
In addition, we claim that
\begin{align}  \label{eq_claim_same_risk_regardless_of_randomization_or_not}
    \max_{\nu_i \in \MM_1^+(\BBB_{\varepsilon}(\xB_i))} \EBB_{\theta \sim \tilde{\mu}, \xB \sim \nu_i} [\ell(\theta, (\xB, y_i))]
    = \EBB_{\theta \sim \tilde{\mu}} [\ell(\theta, (\hat{\xB}_i^*, y_i))],
\end{align}
where $\hat{\xB}_i^* \in \argmax_{\xB \in \BBB_{\varepsilon}(\xB_i)} \EBB_{\theta \sim \tilde{\mu}} [\ell(\theta, (\xB, y_i))]$.
To verify this claim, we first note that 
$
    \max_{\nu_i \in \MM_1^+(\BBB_{\varepsilon}(\xB_i))} \EBB_{\theta \sim \tilde{\mu}, \xB \sim \nu_i} [\ell(\theta, (\xB, y_i))]
    \ge \max_{\xB \in \BBB_{\varepsilon}(\xB_i)} \EBB_{\theta \sim \tilde{\mu}} [\ell(\theta, (\xB, y_i))]
$
because $\delta_{\xB} \in \MM_1^+(\BBB_{\varepsilon}(\xB_i))$ for any $\xB \in \BBB_{\varepsilon}(\xB_i)$.
On the other hand, for some $\nu_i \in \MM_1^+(\BBB_{\varepsilon}(\xB_i))$, we have
\begin{align*}
    \EBB_{\theta \sim \tilde{\mu}, \xB \sim \nu_i} [\ell(\theta, (\xB, y_i))]
    \le \EBB_{\xB \sim \nu_i} \left[ 
       \EBB_{\theta \sim \tilde{\mu}} [\ell(\theta, (\hat{\xB}_i^*, y_i))]
    \right]
    = \EBB_{\theta \sim \tilde{\mu}} [\ell(\theta, (\hat{\xB}_i^*, y_i))].
\end{align*}
Hence, we have proved the claim~\eqref{eq_claim_same_risk_regardless_of_randomization_or_not}.
Combining~\eqref{eq_sup_problem_closed_form} and~\eqref{eq_claim_same_risk_regardless_of_randomization_or_not}, we obtain
\begin{align}  \label{eq_regularization_error_expand}
    &\max_{\nu_i \in \MM_1^+(\BBB_{\varepsilon}(\xB_i))} \EBB_{\theta \sim \tilde{\mu}, \xB \sim \nu_i} [\ell(\theta, (\xB, y_i))]
        - \max_{\nu_i} \left\{\EBB_{\theta \sim \tilde{\mu}, \xB \sim \nu_i} [\ell(\theta, (\xB, y_i))] - \beta \textrm{KL}(\nu_i \| u_i) \right\}  \nonumber \\
    ={}& \EBB_{\theta \sim \tilde{\mu}} [\ell(\theta, (\hat{\xB}_i^*, y_i))]
        - \beta \log \Big( \frac{\int_{\BBB_{\varepsilon}(\xB_i)} \exp(\beta^{-1} \EBB_{\theta \sim \tilde{\mu}}[\ell(\theta, (\xB, y_i))]) d \xB}{\textrm{Vol}(\BBB_{\varepsilon}(\xB_i))}\Big)  \nonumber \\
    ={}& - \beta \log \Big( \frac{\int_{\BBB_{\varepsilon}(\xB_i)} \exp(\beta^{-1} \EBB_{\theta \sim \tilde{\mu}}[\ell(\theta, (\xB, y_i)) - \ell(\theta, (\hat{\xB}_i^*, y_i))]) d \xB}{\textrm{Vol}(\BBB_{\varepsilon}(\xB_i))} \Big)
\end{align}
To proceed, we define $A_{\zeta} \subseteq \BBB_{\varepsilon}(\xB_i)$ as
\begin{align}  \label{eq_A_zeta_def}
    A_{\zeta} = \{\xB \in \BBB_{\varepsilon}(\xB_i) : 
        \EBB_{\theta \sim \tilde{\mu}}[\ell(\theta, (\xB, y_i))] + \zeta
        \ge \EBB_{\theta \sim \tilde{\mu}}[\ell(\theta, (\hat{\xB}_i^*, y_i))]
    \},
\end{align}
where $\zeta > 0$.
Recall that $\BBB_{\varepsilon}(\xB_i) = \{ \xB \in \RBB^{d_{\xB}}: \|\xB - \xB_i\|_{\infty} \le \varepsilon\}$, and that $\ell$ is $G$-Lipschitz continuous w.r.t.\ $\xB$ in terms of the $\ell_{\infty}$ norm.
For any $\xB \in \BBB_{\varepsilon}(\xB_i)$ such that $\|\xB - \hat{\xB}_i^*\|_{\infty} \le \zeta / G$, we have
\begin{align*}
    \EBB_{\theta \sim \tilde{\mu}}[\ell(\theta, (\hat{\xB}_i^*, y_i))]
    - \EBB_{\theta \sim \tilde{\mu}}[\ell(\theta, (\xB, y_i))]
    \le G \| \xB - \hat{\xB}_i^*\|_{\infty} \le \zeta,
\end{align*}
which implies that $\xB \in A_{\zeta}$. 
Thus, the volume of $A_{\zeta}$ satisfies $\textrm{Vol}(A_{\zeta}) \ge \min\{ \varepsilon^{d_{\xB}}, (\zeta / G)^{d_{\xB}}\}$.
Following the argument in~\cite[Section C.6]{meunier2021mixed}, we can bound
\begin{align}  \label{eq_regularization_error_bound}
    &- \beta \log \Big( \frac{\int_{\BBB_{\varepsilon}(\xB_i)} \exp(\beta^{-1} \EBB_{\theta \sim \tilde{\mu}}[\ell(\theta, (\xB, y_i)) - \ell(\theta, (\hat{\xB}_i^*, y_i))]) d \xB}{\textrm{Vol}(\BBB_{\varepsilon}(\xB_i))} \Big)   \nonumber \\
    ={}& - \beta \log \Big(
        \frac{1}{\textrm{Vol}(\BBB_{\varepsilon}(\xB_i))} \int_{A_{\zeta}} \exp(\beta^{-1} \EBB_{\theta \sim \tilde{\mu}}[\ell(\theta, (\xB, y_i)) - \ell(\theta, (\hat{\xB}_i^*, y_i))]) d \xB  \nonumber \\
        &+ \frac{1}{\textrm{Vol}(\BBB_{\varepsilon}(\xB_i))} \int_{\BBB_{\varepsilon}(\xB_i) \backslash A_{\zeta}} \exp(\beta^{-1} \EBB_{\theta \sim \tilde{\mu}}[\ell(\theta, (\xB, y_i)) - \ell(\theta, (\hat{\xB}_i^*, y_i))]) d \xB
    \Big)  \nonumber \\
    \le{}& - \beta \log \frac{\textrm{Vol}(A_{\zeta})}{\exp(\zeta / \beta) \textrm{Vol}(\BBB_{\varepsilon}(\xB_i))}
        - \beta \log \left( 1 + \frac{\exp(\zeta / \beta)}{\textrm{Vol}(A_{\zeta})} \int_{\BBB_{\varepsilon}(\xB_i) \backslash A_{\zeta}} \exp(\beta^{-1} \EBB_{\theta \sim \tilde{\mu}}[\ell(\theta, (\xB, y_i)) - \ell(\theta, (\hat{\xB}_i^*, y_i))]) d \xB \right)  \nonumber \\
    \le{}& \zeta + \beta \log \frac{\textrm{Vol}(\BBB_{\varepsilon}(\xB_i))}{\textrm{Vol}(A_{\zeta})}  \nonumber \\
    \le{}& \zeta + \beta d_{\xB} \max\{ \log 2, \log (2 \varepsilon G / \zeta)\},
\end{align}
where the first inequality follows from~\eqref{eq_A_zeta_def} and the fact that $\log(a + b) = \log a + \log (1 + b / a)$,
and the second inequality follows from the fact that $\log(1 + c) >= 0$ for any $c \ge 0$.
By combining~\eqref{eq_regularization_error_init}, \eqref{eq_regularization_error_expand}, and~\eqref{eq_regularization_error_bound} and optimizing $\zeta$, we obtain
\begin{align*}
    \GM_0(\tilde{\mu}, \tilde{\nu})
    \le
    \begin{cases}
        \GM_{\beta}(\tilde{\mu}, \tilde{\nu}) + \beta d_{\xB} \log 2 + \varepsilon, \quad \textrm{if} \ \varepsilon \le \beta d_{\xB} \\ 
        \GM_{\beta}(\tilde{\mu}, \tilde{\nu}) + \beta d_{\xB} \log \frac{2 \varepsilon G}{\beta d_{\xB}} + \beta d_{\xB}, \quad \textrm{otherwise}.
    \end{cases}
\end{align*}
This concludes the proof.
\end{proof}

\subsection{Proof of Lemma~\ref{lemma_primal_Lyapunov}}
\begin{proof}
The proof follows from~\cite[Lemmas 8 and 9]{domingo2020mean}.
Specifically, following the argument of Eq.\ (34) in~\cite[Lemma 8]{domingo2020mean}, the following inequality holds under Assumptions~\ref{assumption_loss_function}-\ref{assumption_initial_value}
\begin{align*}
    \frac{1}{T} \int_0^T \LM(\mu(t), \nu(t)) - \LM(\mu^*(T), \bar{\nu}(T))
    \le (U + G) \inf_{\mu \in \MM_1^+(\Theta)} \left\{ 
        \|\mu^*(T) - \mu\|_{\textrm{BL}}^* + \frac{1}{\alpha T (U + G)} \textrm{KL}(\mu, \mu^0)
        \right\}
        + \frac{\gamma (U + G)^2 T}{2},
\end{align*}
where $\|\cdot\|_{\textrm{BL}}^*$ denotes the dual of the bounded Lipschitz norm.
By Lemma 9 in~\cite{domingo2020mean}, we have the following bound 
\begin{align*}
    \inf_{\mu \in \MM_1^+(\Theta)} \left\{ 
        \|\mu^*(T) - \mu\|_{\textrm{BL}}^* + \frac{1}{\alpha T} \textrm{KL}(\mu, \mu^0)
        \right\}
    \le \frac{d_{\theta}}{\alpha T (U + G)} ( 1 - \log d_{\theta} + \log ( \alpha T (U + G))) + \frac{K + K'}{\alpha T (U + G)}.
\end{align*}
Combining the above two inequalities leads to the desired result.
\end{proof}

\subsection{Proof of Lemma~\ref{lemma_dual_Lyapunov}}
Let us first recall the definition of Gateaux derivation, which will be used in the proof of Lemma~\ref{lemma_dual_Lyapunov}.
\begin{definition}
On Banach spaces $X$ and $Y$, the Gateaux derivative of a function $F(\cdot): X \rightarrow Y$ at the point $\xB \in X$ in the direction $\uB \in X$ is defined as
\begin{equation}  \label{eq_Gateaux}
    D F(\xB) \uB = \lim_{\delta \downarrow 0} \frac{1}{\delta} \left( F(\xB + \delta \uB) - F(\xB) \right).
\end{equation}
\end{definition}

Now, we are ready to prove Lemma~\ref{lemma_dual_Lyapunov}.
\begin{proof}
We mainly follow the proof technique of~\cite[Proposition D.4]{perkins2014stochastic} in which both the players in the zero-sum game follow the LBR flow.
In contrast, in our WFR-LRB flow, only the classifier follows the LBR flow.
In addition, we construct and analyze the potential function $\RM_{\nu}(T)$ in~\eqref{eq_Lyapunov_functions} which is different from theirs.
Notice that when $t > 1$, the derivative of $\bar{\nu}$ is given by
\begin{align}  \label{eq_derivative_nu_bar}
    \dot{\bar{\nu}}(t)
    ={}& \lim_{\delta \downarrow 0} \frac{\bar{\nu}(t + \delta) - \bar{\nu}(t)}{\delta}  \nonumber \\
    ={}& \lim_{\delta \downarrow 0} \frac{\frac{1}{t + \delta} \int_0^{t + \delta} \nu(s) d s - \frac{1}{t} \int_0^t \nu(s) d s}{\delta}  \nonumber \\
    ={}& \lim_{\delta \downarrow 0} \big\{ \frac{1}{\delta}(\frac{1}{t \!+\! \delta} \!-\! \frac{1}{t}) \! \int_0^t \! \nu(s) d s \!+\! \frac{1}{(t \!+\! \delta) \delta} \! \int_t^{t \!+\! \delta} \! \nu(s) d s \big\} \nonumber \\
    ={}& - \frac{1}{t^2} \int_0^t \nu(s) d s + \frac{1}{t} \nu(t)  \nonumber \\
    ={}& \frac{1}{t} \left( \nu(t) - \bar{\nu}(t) \right)
\end{align}
For $\mu \in \MM_1^+(\Theta)$ and $\nu \in \Sigma$, we define
$\mu \otimes \nu := (\mu \otimes \nu_1, \cdots \mu \otimes \nu_N) \in \MM_1^+(\Theta \times \BBB_{\varepsilon}(\xB_1)) \times \ldots \times \MM_1^+(\Theta \times \BBB_{\varepsilon}(\xB_N))$,
where $\mu \otimes \nu_i$ denotes the product measure of $\mu$ and $\nu_i$ on $\Theta \times \BBB_{\varepsilon}(\xB_i)$.
We let $t = e^{\widetilde{t}}$ for $\widetilde{t} \in [-\infty, \infty)$ with the convention $e^{-\infty} = 0$ and define
\begin{align}
\begin{cases}
    \breve{\mu}(\widetilde{t}) = \bar{\mu}(e^{\widetilde{t}}) \\
    \breve{\nu}(\widetilde{t}) = \bar{\nu}(e^{\widetilde{t}}),
\end{cases}
\end{align}
where $\bar{\mu}(t)$ and $\bar{\nu}(t)$ are defined in Section~\ref{section_algorithm}.
Note that $\breve{\mu}(\widetilde{t})$ and $\breve{\nu}(\widetilde{t})$ satisfy
\begin{align}
\begin{cases}
    \dot{\breve{\mu}}(\widetilde{t}) = \widetilde{\mu}(\widetilde{t}) - \breve{\mu}(\widetilde{t}) \\
    \dot{\breve{\nu}}(\widetilde{t}) = \widetilde{\nu}(\widetilde{t}) - \breve{\nu}(\widetilde{t}),
\end{cases}
\end{align}
with initial conditions $\breve{\mu}(0) = \bar{\mu}(1)$ and $\breve{\nu}(0) = \nu^{0}$, where $\widetilde{\mu}(\widetilde{t}) = \mu(e^{\widetilde{t}})$ and $\widetilde{\nu}(\widetilde{t}) = \nu(e^{\widetilde{t}})$.
Let $\zeta(t) := \mu(t) \otimes \nu(t)$ and $\bar{\zeta}(t) := \frac{1}{t} (\int_0^t \zeta(s) d s)$ for $t \in [1, \infty)$.
Following the same argument as~\eqref{eq_derivative_nu_bar}, $\bar{\zeta}(t)$ satisfies the following differential equation
\begin{align}
    \dot{\bar{\zeta}}(t) = \frac{1}{t} (\zeta(t) - \bar{\zeta}(t))
\end{align}
with initial condition $\bar{\zeta}(1) = \int_{t=0}^1 \mu(t) \otimes \nu^{0} d t = \bar{\mu}(1) \otimes \nu^{0}$.
Note that $\bar{\zeta}(t)$ can be reparameterized as
\begin{align}
    \breve{\zeta}(\widetilde{t}) = \int_{-\infty}^{\widetilde{t}} e^{\widetilde{s} - \widetilde{t}} \ \widetilde{\zeta}(\widetilde{s}) d \widetilde{s}.
\end{align}
One can easily show that $\breve{\zeta}(\widetilde{t})$ satisfies the following differential equation
\begin{align}
    \dot{\breve{\zeta}}(\widetilde{t}) = \widetilde{\zeta}(\widetilde{t}) - \breve{\zeta}(\widetilde{t})
\end{align}
with initial condition $\breve{\zeta}(0) = \breve{\mu}(0) \otimes \nu^{0}$.
We let $T = e^{\widetilde{T}}$ and reparameterize $\RM_{\nu}(T)$ in~\eqref{eq_Lyapunov_functions} as
\begin{align}  \label{eq_R_tilde_nu_def}
    \widetilde{\RM}_{\nu}(\widetilde{T})
    :={}& \RM_{\nu}(T)
    = \LM(\bar{\mu}(T), \nu(T))
        - \beta \HM(\nu(T))
        - \frac{1}{T} \int_0^T \LM(\mu(t), \nu(t))
        + \beta \HM(\bar{\nu}(T)),  \nonumber \\
    ={}& \LM(\bar{\mu}(T), \nu(T))
        - \beta \HM(\nu(T))
        - \frac{1}{N} \sum_{i=1}^N \EBB_{(\theta, \xB) \sim \bar{\zeta}_i(T)}[\ell(\theta, \xB)]
        + \beta \HM(\bar{\nu}(T)),  \nonumber \\
    ={}& \LM(\breve{\mu}(\widetilde{T}), \widetilde{\nu}(\widetilde{T}))
        - \beta \HM(\widetilde{\nu}(\widetilde{T}))
        - \frac{1}{N} \sum_{i=1}^N \EBB_{(\theta, \xB) \sim \breve{\zeta}_i(\widetilde{T})}[\ell(\theta, \xB)]
        + \beta \HM(\breve{\nu}(\widetilde{T})),
\end{align}
where the second equality holds because
\begin{align}
    \frac{1}{T} \int_0^T \LM(\mu(t), \nu(t))
    ={}& \frac{1}{N} \sum_{i=1}^N \frac{1}{T} \int_0^T
        \int_{\Theta} \int_{\BBB_{\varepsilon}(\xB_i)}
            \ell(\theta, (\xB, y_i))
        \nu_i(t) (d \xB) \mu(t) (d \theta)  \nonumber \\
    ={}& \frac{1}{N} \sum_{i=1}^N \frac{1}{T} \int_0^T
        \int_{\Theta \times \BBB_{\varepsilon}(\xB_i)}
            \ell(\theta, (\xB, y_i))
        (\mu(t) \otimes \nu_i(t)) (d (\theta, \xB))  \nonumber \\
    ={}& \frac{1}{N} \sum_{i=1}^N
        \int_{\Theta \times \BBB_{\varepsilon}(\xB_i)}
            \ell(\theta, (\xB, y_i))
        \frac{1}{T} \int_0^T \zeta_i(t) (d (\theta, \xB))  \nonumber \\
    ={}& \frac{1}{N} \sum_{i=1}^N
        \int_{\Theta \times \BBB_{\varepsilon}(\xB_i)}
            \ell(\theta, (\xB, y_i))
        \bar{\zeta}_i(T) (d (\theta, \xB)).
\end{align}
In what follows, we provide an upper bound of $\widetilde{\RM}_{\nu}(\widetilde{T})$.
Notice that when $\widetilde{T} = 0$,
\begin{align}
    \widetilde{\RM}_{\nu}(0)
    ={}& \LM(\breve{\mu}(0), \widetilde{\nu}(0)) - \beta \HM(\widetilde{\nu}(0)) 
        - \Big(\LM(\breve{\mu}(0), \nu^{0}) - \beta \HM(\nu^{0}) \Big)  \nonumber \\
    ={}& \max_{\nu \in \Sigma}
        \left\{
            \LM(\breve{\mu}(0), \nu) - \beta \HM(\nu)
        \right\}
        - \Big(\LM(\breve{\mu}(0), \nu^{0}) - \beta \HM(\nu^{0}) \Big)
    \ge 0.
\end{align}
Let $\widetilde{q}_i(\widetilde{t}, \xB)$ denote the probability density of the distribution $\widetilde{\nu}_i(\widetilde{t})$ at $\xB \in \XM$.
By~\eqref{eq_nu_t} and~\eqref{eq_density_logit_best_response}, we have
\begin{align}  \label{eq_density_best_fixed}
    \widetilde{q}_i(\widetilde{t}, \xB) = \frac{\exp(\beta^{-1} \EBB_{\theta \sim \breve{\mu}(\widetilde{t})}[\ell(\theta, (\xB, y_i))])}{\int_{\BBB_{\varepsilon}(\xB_i)} \exp(\beta^{-1} \EBB_{\theta \sim \breve{\mu}(\widetilde{t})}[\ell(\theta, (\xB, y_i))]) d \xB}.
\end{align}
We can rewrite $\beta \HM(\widetilde{\nu}(\widetilde{t}))$ as
\begin{align}  \label{eq_regularizer_rewritten}
   \beta \HM(\widetilde{\nu}(\widetilde{t}))
   ={}& \frac{\beta}{N} \sum_{i=1}^N \int_{\BBB_{\varepsilon}(\xB_i)} \log(\widetilde{q}_i(\widetilde{t}, \xB)) \widetilde{q}_i(\widetilde{t}, \xB) d \xB
        + \frac{\beta}{N} \sum_{i=1}^N \log(\textrm{Vol}(\BBB_{\varepsilon}(\xB_i)))  \nonumber \\
    ={}& \frac{\beta}{N} \sum_{i=1}^N \int_{\BBB_{\varepsilon}(\xB_i)} \log(\widetilde{q}_i(\widetilde{t}, \xB)) \widetilde{q}_i(\widetilde{t}, \xB) d \xB_i
        + \frac{\beta}{N} \sum_{i=1}^N \log(\textrm{Vol}(\BBB_{\varepsilon}(\xB_i)))  \nonumber \\
    ={}& \frac{\beta}{N} \sum_{i=1}^N \int_{\BBB_{\varepsilon}(\xB_i)} \beta^{-1} \EBB_{\theta \sim \breve{\mu}(\widetilde{t})}[\ell(\theta, (\xB, y_i))] \widetilde{q}_i(\widetilde{t}, \xB) d \xB
        + \frac{\beta}{N} \sum_{i=1}^N \log(\textrm{Vol}(\BBB_{\varepsilon}(\xB_i)))  \nonumber \\ 
        &- \frac{\beta}{N} \sum_{i=1}^N \log \left( \int_{\BBB_{\varepsilon}(\xB_i)} \exp(\beta^{-1} \EBB_{\theta \sim \breve{\mu}(\widetilde{t})}[\ell(\theta, (\xB, y_i))]) d \xB \right)  \nonumber \\
    ={}& \frac{1}{N} \sum_{i=1}^N \EBB_{\theta \sim \breve{\mu}(\widetilde{t}), \xB \sim \widetilde{\nu}_i(\widetilde{t})}[\ell(\theta, (\xB, y_i))]
        + \frac{\beta}{N} \sum_{i=1}^N \log(\textrm{Vol}(\BBB_{\varepsilon}(\xB_i)))  \nonumber \\
        &- \frac{\beta}{N} \sum_{i=1}^N \log \left( \int_{\BBB_{\varepsilon}(\xB_i)} \exp(\beta^{-1} \EBB_{\theta \sim \breve{\mu}(\widetilde{t})}[\ell(\theta, (\xB, y_i))]) d \xB \right)  \nonumber \\
    ={}& \LM(\breve{\mu}(\widetilde{t}), \widetilde{\nu}(\widetilde{t}))
        + \frac{\beta}{N} \sum_{i=1}^N \log(\textrm{Vol}(\BBB_{\varepsilon}(\xB_i)))  \nonumber \\
        &- \frac{\beta}{N} \sum_{i=1}^N \log \left( \int_{\BBB_{\varepsilon}(\xB_i)} \exp(\beta^{-1} \EBB_{\theta \sim \breve{\mu}(\widetilde{t})}[\ell(\theta, (\xB, y_i))]) d \xB \right).
\end{align}
Plugging~\eqref{eq_regularizer_rewritten} into the definition of $\widetilde{\RM}_{\nu}(\widetilde{t})$ in~\eqref{eq_R_tilde_nu_def}, we obtain
\begin{align}  \label{eq_dual_Lyapunov_rewritten}
    \widetilde{\RM}_{\nu}(\widetilde{t})
    ={}& \frac{\beta}{N} \sum_{i=1}^N \log \left( \int_{\BBB_{\varepsilon}(\xB_i)} \exp(\beta^{-1} \EBB_{\theta \sim \breve{\mu}(\widetilde{t})}[\ell(\theta, (\xB, y_i))]) d \xB \right)
        - \frac{\beta}{N} \sum_{i=1}^N \log(\textrm{Vol}(\BBB_{\varepsilon}(\xB_i)))  \nonumber \\
        &- \frac{1}{N} \sum_{i=1}^N \EBB_{(\theta, \xB) \sim \breve{\zeta}_i(\widetilde{t})}[\ell(\theta, (\xB, y_i))]
        + \beta \HM(\breve{\nu}(\widetilde{t})).
\end{align}
Taking the derivative of $\widetilde{\RM}_{\nu}(\widetilde{t})$ w.r.t.\ $\widetilde{t}$ and recalling the definition of Gateaux derivative~\eqref{eq_Gateaux}, we obtain
\begin{align}  \label{eq_derivative_dual_Lyapunov}
    \dot{\widetilde{\RM}}_{\nu}(\widetilde{t})
    ={}& \frac{\beta}{N} \sum_{i=1}^N D \log \left( \int_{\BBB_{\varepsilon}(\xB_i)} \exp(\beta^{-1} \EBB_{\theta \sim \breve{\mu}(\widetilde{t})}[\ell(\theta, (\xB, y_i))]) d \xB \right) \dot{\breve{\mu}}(\widetilde{t})  \nonumber \\
        &+ \beta D \HM(\breve{\nu}(\widetilde{t})) \dot{\breve{\nu}}(\widetilde{t})
        - \frac{1}{N} \sum_{i=1}^N D \EBB_{(\theta, \xB) \sim \breve{\zeta}_i(\widetilde{t})}[\ell(\theta, \xB)] \dot{\breve{\zeta}}_i(\widetilde{t}).
\end{align}
The first term on the RHS of~\eqref{eq_derivative_dual_Lyapunov} can be written as
\begin{align}  \label{eq_derivative_dual_Lyapunov_1st_term}
    &\frac{\beta}{N} \sum_{i=1}^N D \log \left( \int_{\BBB_{\varepsilon}(\xB_i)} \exp \left(\beta^{-1} \EBB_{\theta \sim \breve{\mu}(\widetilde{t})}[\ell(\theta, (\xB, y_i))] \right) d \xB \right) \dot{\breve{\mu}}(\widetilde{t}) \\
    ={}& \frac{\beta}{N} \sum_{i=1}^N \frac{1}{\int_{\BBB_{\varepsilon}(\xB_i)} \exp(\beta^{-1} \EBB_{\theta \sim \breve{\mu}(\widetilde{t})}[\ell(\theta, (\xB, y_i))]) d \xB}
        \int_{\BBB_{\varepsilon}(\xB_i)} D \exp \left( \beta^{-1} \EBB_{\theta \sim \breve{\mu}(\widetilde{t})}[\ell(\theta, (\xB, y_i))]\right) \dot{\breve{\mu}}(\widetilde{t}) d \xB  \nonumber \\
    ={}& \frac{\beta}{N} \sum_{i=1}^N 
    \int_{\BBB_{\varepsilon}(\xB_i)} 
        \beta^{-1} D \EBB_{\theta \sim \breve{\mu}(\widetilde{t})}[\ell(\theta, (\xB, y_i))] \dot{\breve{\mu}}(\widetilde{t})
        \frac{\exp \left( \beta^{-1} \EBB_{\theta \sim \breve{\mu}(\widetilde{t})}[\ell(\theta, (\xB, y_i))]\right)}{\int_{\BBB_{\varepsilon}(\xB_i)} \exp(\beta^{-1} \EBB_{\theta \sim \breve{\mu}(\widetilde{t})}[\ell(\theta, (\xB, y_i))]) d \xB}
    d \xB  \nonumber \\
    \overset{(a)}{=}{}& \frac{1}{N} \sum_{i=1}^N 
    \int_{\BBB_{\varepsilon}(\xB_i)} 
        \widetilde{q}_i(\widetilde{t})
        \int_{\Theta} \ell(\theta, (\xB, y_i)) D \breve{\mu}(\widetilde{t}) \dot{\breve{\mu}}(\widetilde{t}) (d \theta)
    d \xB  \nonumber \\
    ={}& \frac{1}{N} \sum_{i=1}^N 
    \int_{\BBB_{\varepsilon}(\xB_i)} 
        \widetilde{q}_i(\widetilde{t})
        \int_{\Theta}[\ell(\theta, (\xB, y_i))] \dot{\breve{\mu}}(\widetilde{t}) (d \theta)
    d \xB  \nonumber \\
    ={}& \LM(\dot{\breve{\mu}}(\widetilde{t}), \widetilde{\nu}(\widetilde{t})),
\end{align}
where (a) holds because $D \breve{\mu}(\widetilde{t}) \dot{\breve{\mu}}(\widetilde{t}) = \dot{\breve{\mu}}(\widetilde{t})$~\citep{lahkar2014continuous}.
To proceed, we denote $\breve{q}_i(\widetilde{t}, \xB)$ as the density of $\breve{\nu}_i(\widetilde{t})$ at $\xB$.
Then, we rewrite the second term on the RHS of~\eqref{eq_derivative_dual_Lyapunov} as
\begin{align}  \label{eq_derivative_dual_Lyapunov_2nd_term}
    \beta D \HM(\breve{\nu}(\widetilde{t})) \dot{\breve{\nu}}(\widetilde{t})
    ={}& \frac{\beta}{N} \sum_{i=1}^N D \left( \textrm{KL}(\breve{\nu}_i(\widetilde{t}) \| u_i) \right) \dot{\breve{\nu}}_i(\widetilde{t})  \nonumber \\
    ={}& \frac{\beta}{N} \sum_{i=1}^N D \left( \int_{\BBB_{\varepsilon}(\xB_i)} \log \breve{q}_i(\widetilde{t}, \xB) \ \breve{\nu}_i(\widetilde{t}) (d \xB) \right) \dot{\breve{\nu}}_i(\widetilde{t})  \nonumber \\
    ={}& \frac{\beta}{N} \sum_{i=1}^N  \left\{
        \int_{\BBB_{\varepsilon}(\xB_i)} D \left( \log \breve{q}_i(\widetilde{t}, \xB) \right) \dot{\breve{\nu}}_i(\widetilde{t}) \ \breve{\nu}_i(\widetilde{t}) (d \xB)
        + \int_{\BBB_{\varepsilon}(\xB_i)} \log \breve{q}_i(\widetilde{t}, \xB) D \breve{\nu}_i(t) \dot{\breve{\nu}}_i(\widetilde{t}) (d \xB)
    \right\}  \nonumber \\
    ={}& \frac{\beta}{N} \sum_{i=1}^N  \left\{
        \int_{\BBB_{\varepsilon}(\xB_i)} \frac{1}{\breve{q}_i(\widetilde{t}, \xB)} D \breve{q}_i(\widetilde{t}, \xB) \dot{\breve{\nu}}_i(\widetilde{t}) \ \breve{q}_i(\widetilde{t}, \xB) (d \xB)
        + \int_{\BBB_{\varepsilon}(\xB_i)} \log \breve{q}_i(\widetilde{t}, \xB) \dot{\breve{\nu}}_i(\widetilde{t}) (d \xB)
    \right\}  \nonumber \\
    ={}& \frac{\beta}{N} \sum_{i=1}^N  \left\{
        \int_{\BBB_{\varepsilon}(\xB_i)} \dot{\breve{\nu}}_i(\widetilde{t}) (d \xB)
        + \int_{\BBB_{\varepsilon}(\xB_i)} \log \breve{q}_i(\widetilde{t}, \xB) \dot{\breve{\nu}}_i(\widetilde{t}) (d \xB)
    \right\}  \nonumber \\
    ={}& \frac{\beta}{N} \sum_{i=1}^N
        \int_{\BBB_{\varepsilon}(\xB_i)} \log \breve{q}_i(\widetilde{t}, \xB) \dot{\breve{\nu}}_i(\widetilde{t}) (d \xB).
\end{align}
The third term on the RHS of~\eqref{eq_derivative_dual_Lyapunov} can be rewritten as
\begin{align}  \label{eq_derivative_dual_Lyapunov_3rd_term}
    - \frac{1}{N} \sum_{i=1}^N D \EBB_{(\theta, \xB) \sim \breve{\zeta}_i(\widetilde{t})}[\ell(\theta, \xB)] \dot{\breve{\zeta}}_i(\widetilde{t})
    ={}& - \frac{1}{N} \sum_{i=1}^N  \int_{\Theta \times \BBB_{\varepsilon}(\xB_i)} \ell(\theta, (\xB, y_i)) \left( D \breve{\zeta}_i(\widetilde{t}) \dot{\breve{\zeta}}_i(\widetilde{t}) \right) (d (\theta, \xB))  \nonumber \\
    ={}& - \frac{1}{N} \sum_{i=1}^N  \int_{\Theta \times \BBB_{\varepsilon}(\xB_i)} \ell(\theta, (\xB, y_i)) \left( \dot{\breve{\zeta}}_i(\widetilde{t}) (d (\theta, \xB)) \right).
\end{align}
Combining~\eqref{eq_derivative_dual_Lyapunov}, \eqref{eq_derivative_dual_Lyapunov_1st_term}, \eqref{eq_derivative_dual_Lyapunov_2nd_term}, and~\eqref{eq_derivative_dual_Lyapunov_3rd_term}, we arrive at
\begin{align}  \label{eq_derivative_dual_Lyapunov_convergence}
    \dot{\widetilde{\RM}}_{\nu}(\widetilde{t})
    ={}& \LM(\dot{\breve{\mu}}(\widetilde{t}), \widetilde{\nu}(\widetilde{t}))
        + \frac{\beta}{N} \sum_{i=1}^N \int_{\BBB_{\varepsilon}(\xB_i)} \log \breve{q}_i(\widetilde{t}, \xB) \dot{\breve{\nu}}_i(\widetilde{t}) (d \xB)  \nonumber \\
        &- \frac{1}{N} \sum_{i=1}^N  \int_{\Theta \times \BBB_{\varepsilon}(\xB_i)} \ell(\theta, (\xB, y_i)) \left( \dot{\breve{\zeta}}_i(\widetilde{t}) (d (\theta, \xB)) \right)  \nonumber \\
    ={}& \LM(\widetilde{\mu}(\widetilde{t}) - \breve{\mu}(\widetilde{t}), \widetilde{\nu}(\widetilde{t}))
        + \frac{\beta}{N} \sum_{i=1}^N \int_{\BBB_{\varepsilon}(\xB_i)} \log \breve{q}_i(\widetilde{t}, \xB) \left(\widetilde{\nu}_i(\widetilde{t}) \right) (d \xB)
        - \frac{\beta}{N} \sum_{i=1}^N \int_{\BBB_{\varepsilon}(\xB_i)} \log \breve{q}_i(\widetilde{t}, \xB) \left( \breve{\nu}_i(\widetilde{t}) \right) (d \xB)  \nonumber \\
        &- \frac{1}{N} \sum_{i=1}^N  \int_{\Theta \times \BBB_{\varepsilon}(\xB_i)} \ell(\theta, (\xB, y_i)) \left( \tilde{\zeta}_i(\widetilde{t}) (d (\theta, \xB)) \right)
        + \frac{1}{N} \sum_{i=1}^N  \int_{\Theta \times \BBB_{\varepsilon}(\xB_i)} \ell(\theta, (\xB, y_i)) \left( \breve{\zeta}_i(\widetilde{t}) (d (\theta, \xB)) \right)  \nonumber \\
    ={}& \LM(\widetilde{\mu}(\widetilde{t}), \widetilde{\nu}(\widetilde{t}))
        - \LM(\breve{\mu}(\widetilde{t}), \widetilde{\nu}(\widetilde{t}))
        - \LM(\widetilde{\mu}(\widetilde{t}), \widetilde{\nu}(\widetilde{t}))
        + \frac{1}{N} \sum_{i=1}^N \EBB_{(\theta, \xB) \sim \breve{\zeta}_i(\widetilde{t})}[\ell(\theta, \xB)]  \nonumber \\
        &+ \frac{\beta}{N} \sum_{i=1}^N \int_{\BBB_{\varepsilon}(\xB_i)}
            \left( \log \breve{q}_i(\widetilde{t}, \xB)
                - \log \widetilde{q}_i(\widetilde{t}, \xB)
                + \log \widetilde{q}_i(\widetilde{t}, \xB)
                - \log(\textrm{Vol}(\BBB_{\varepsilon}(\xB_i)))
            \right)
        \left(\widetilde{\nu}_i(\widetilde{t}) \right) (d \xB)  \nonumber \\
        &+ \frac{\beta}{N} \sum_{i=1}^N \int_{\BBB_{\varepsilon}(\xB_i)} 
            \left( \log(\textrm{Vol}(\BBB_{\varepsilon}(\xB_i)))
                - \log \breve{q}_i(\widetilde{t}, \xB)
            \right)
        \left( \breve{\nu}_i(\widetilde{t}) \right) (d \xB)  \nonumber \\
    ={}& \LM(\widetilde{\mu}(\widetilde{t}), \widetilde{\nu}(\widetilde{t}))
        - \LM(\breve{\mu}(\widetilde{t}), \widetilde{\nu}(\widetilde{t}))
        - \LM(\widetilde{\mu}(\widetilde{t}), \widetilde{\nu}(\widetilde{t}))
        + \frac{1}{N} \sum_{i=1}^N \EBB_{(\theta, \xB) \sim \breve{\zeta}_i(\widetilde{t})}[\ell(\theta, \xB)]  \nonumber \\
        &+ \frac{\beta}{N} \sum_{i=1}^N \left(
            - \textrm{KL}(\widetilde{\nu}_i(\widetilde{t}) \| \breve{\nu}_i(\widetilde{t}))
            + \textrm{KL}(\widetilde{\nu}_i(\widetilde{t}) \| u_i)
            - \textrm{KL}(\breve{\nu}_i(\widetilde{t}) \| u_i)
        \right)  \nonumber \\
    ={}& - \left( \LM(\breve{\mu}(\widetilde{t}), \widetilde{\nu}(\widetilde{t}))
        - \beta \HM(\widetilde{\nu}(\widetilde{t}))
        - \frac{1}{N} \sum_{i=1}^N \EBB_{(\theta, \xB) \sim \breve{\zeta}_i(\widetilde{t})}[\ell(\theta, \xB)]
        + \beta \HM(\breve{\nu}(\widetilde{t})) \right)
        - \frac{\beta}{N} \sum_{i=1}^N
            \textrm{KL}(\widetilde{\nu}_i(\widetilde{t}) \| \breve{\nu}_i(\widetilde{t}))
        \nonumber \\
    ={}& - \widetilde{\RM}_{\nu}(\widetilde{t})
        - \frac{\beta}{N} \sum_{i=1}^N
            \textrm{KL}(\widetilde{\nu}_i(\widetilde{t}) \| \breve{\nu}_i(\widetilde{t}))  \nonumber \\
    \le{}& - \widetilde{\RM}_{\nu}(\widetilde{t}),
\end{align}
where the inequality follows from the fact that the KL divergence between any two probability measures is non-negative.
Applying Gronwall's inequality to~\eqref{eq_derivative_dual_Lyapunov_convergence} yields
\begin{align}
    \widetilde{\RM}_{\nu}(\widetilde{t}) \le \widetilde{\RM}_{\nu}(0) e^{- \widetilde{t}}, \quad \forall \widetilde{t} \ge 0.
\end{align}
Recall that $\widetilde{\RM}_{\nu}(\widetilde{t}) = \RM_{\nu}(t)$ with $t = e^{\widetilde{t}}$.
Thus, we obtain
\begin{align}
    \RM_{\nu}(t) \le \frac{\RM_{\nu}(1)}{t}, \quad \forall t \ge 1.
\end{align}
\end{proof}

\section{Details of The Experimental Setting on CIFAR-10 and CIFAR-100}  \label{section_exp_setting}
Following~\cite{meunier2021mixed}, we use ResNet18~\cite{he2016deep} as the base classification model and define the loss function as the TRADES loss~\cite{zhang2019theoretically}.
In the training process, we use the momentum SGD with a minibatch size $1024$ to update the classification models, where the learning rate is initialized as $0.4$ and divided by $10$ once the performance does not improve in the last $10$ epochs.
We set the momentum and weight decay parameters to $0.9$ and $5 \times 10^{-4}$, and set the performance metric of the learning rate scheduler as the robust test accuracy against the PGD\textsubscript{$20$} attack.
In each iteration of SAT and ATM, we use PGD\textsubscript{10} to generate perturbed data samples.
The number of steps in APGD\textsubscript{CE} and APGD\textsubscript{DLR} attacks is set to $100$.
All experiments were implemented in Pytorch and run on a workstations with $2$ Intel E5-2680 v4 CPUs ($28$ cores), $4$ NVIDIA RTX 2080Ti GPUs, and $378$GB memory.

\end{document}